\documentclass[10pt,twocolumn,letterpaper]{article}

\usepackage[pagenumbers]{cvpr} %
\usepackage{times}
\usepackage{epsfig}
\usepackage{graphicx}
\usepackage{amsmath}
\usepackage{amssymb}
\usepackage{color}
\usepackage[dvipsnames]{xcolor}

\usepackage{multirow}
\usepackage{multicol}
\usepackage{enumitem}
\usepackage{booktabs}
\usepackage{caption}
\usepackage{array}
\usepackage[pagebackref,breaklinks,colorlinks,citecolor=cvprblue]{hyperref}
\usepackage[capitalize]{cleveref}
\crefname{section}{Sec.}{Secs.}
\Crefname{section}{Section}{Sections}
\Crefname{table}{Table}{Tables}
\crefname{table}{Tab.}{Tabs.}

\def\eg{\textit{e.g.}\@\xspace}
\def\ie{\textit{i.e.}\@\xspace}

\def\nointeraction{\texttt{no\_interaction}\@\xspace}

\definecolor{cvprblue}{rgb}{0.21,0.49,0.74}

\def\paperID{6084} %
\def\confName{CVPR}
\def\confYear{2024}

\title{Diagnosing Human-object Interaction Detectors}
\begin{document}
\author{Fangrui Zhu\textsuperscript{\rm 1}, \quad Yiming Xie\textsuperscript{\rm 1}, 
\quad Weidi Xie\textsuperscript{\rm 2}, \quad  Huaizu Jiang\textsuperscript{\rm 1} \\
\textsuperscript{\rm 1}Northeastern University~~~~~~\textsuperscript{\rm2} Shanghai Jiao Tong University\\
{\tt\small \{zhu.fang,xie.yim,h.jiang\}@northeastern.edu}, {\tt\small weidi@sjtu.edu.cn}
}

\maketitle

\begin{abstract}
We have witnessed significant progress in human-object interaction (HOI) detection.
The reliance on $mAP$ (mean Average Precision) scores as a summary metric, however, does not provide sufficient insight into the nuances of model performance (\eg, why one model is better than another), which can hinder further innovation in this field.
To address this issue, 
in this paper, we introduce a diagnosis toolbox to provide detailed quantitative break-down analysis of HOI detection models, inspired by the success of object detection diagnosis toolboxes.
We first conduct holistic investigations in the pipeline of HOI detection.
By defining a set of errors and the oracles to fix each of them, we can have a quantitative analysis of the significance of different errors according to the $mAP$ improvement obtained from fixing each error.
We then delve into two sub-tasks of HOI detection:  human-object pair detection and interaction classification, respectively.
For the first detection task, 
we compute the coverage of ground-truth human-object pairs as well as the noisiness level in the detection results.
For the second classification task, we measure a model's performance of differentiating positive and negative detection results
and also classifying the actual interactions when the human-object pairs are correctly detected.
We analyze eight state-of-the-art HOI detection models and provide valuable diagnosis insights to foster future research.
For instance, our diagnosis shows that state-of-the-art model RLIPv2 outperforms others mainly because it significantly improves the multi-label interaction classification accuracy.
Our toolbox is applicable for different methods across different datasets and available at \url{https://github.com/neu-vi/Diag-HOI}.
\end{abstract}

\section{Introduction}
\begin{figure}[t]
    \centering
    \includegraphics[width=\linewidth]{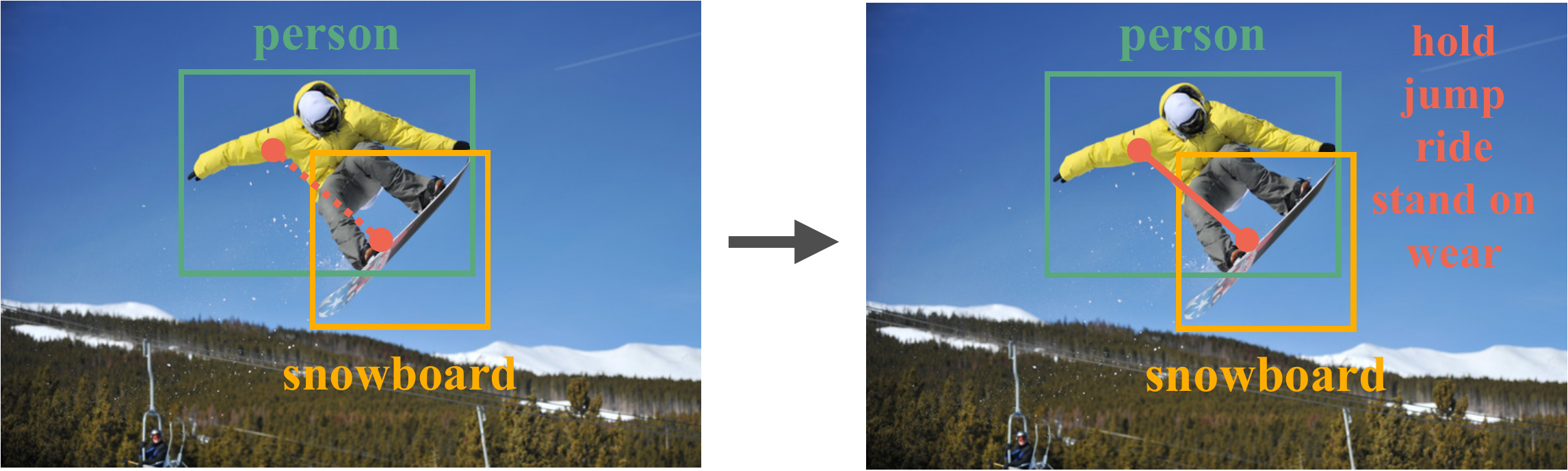}
    \caption{\textbf{Illustration of two sub-tasks in HOI detection.} (a) Detect all human-object pairs that have interactions~(\texttt{person} and \texttt{snowboard}). (b) Classify the interactions between them~(\texttt{hold}, \texttt{jump}, \texttt{ride}, \texttt{stand on}, and \texttt{wear}). }
    \label{fig:problem_describ}
\end{figure}

Human-object interaction~(HOI) detection aims to jointly detect the humans and objects that have interactions in static images.
For example, the person and snowboard in Fig.~\ref{fig:problem_describ}.
\textrm{It provides structured interpretations of the semantics of visual scenes rather than just object recognition or detection.}
A successful HOI detection system is an essential building block for many downstream applications, such as visual question answering~\cite{antol2015vqa,anderson2018bottom,shih2016look,lu2016hierarchical,wang2017fvqa,lu2016hierarchical}, image captioning~\cite{vinyals2016show,aneja2018convolutional,feng2019unsupervised,li2019entangled} and retrieval~\cite{chao2015hico,brown2020smooth,ng2020solar,teichmann2019detect,radenovic2018fine}, etc.

Recent advancements in HOI detection have been marked by increasing mean Average Precision ($mAP$) scores across standard benchmarks~\cite{gupta2015visual,chao2018learning,gao2020drg,ulutan2020vsgnet,gupta2019no,zhou2019relation,zhang2021mining,zhang2021spatially,zhang2022exploring,liao2022gen,yuanrlip,yu2023fine,ma2023fgahoi,li2022discovering,wu2022mining,zhong2022towards,jiang2022bongard,liu2022interactiveness,kim2023relational,yuan2023rlipv2}, denoting remarkable progress.
Nonetheless, the reliance on $mAP$ scores as a summary metric does not provide sufficient insight into the nuances of model performance, including the factors making one method perform better than another or any bottleneck for further improvement.
This lack of detailed understanding may impede future advancements in the field.
The same issue also existed in object detection, a sub task of HOI detection, where $mAP$ is also the dominant evaluation metric.
To address it, diagnosis toolboxes have been designed to provide more useful quantitative break-down analysis~\cite{hoiem2012diagnosing,bolya2020tide}, which have significantly boosted the development of object detection.

In this paper, we aim to replicate the success of these work by introducing a toolbox designed for HOI detection, fostering future research.
Generally speaking, the HOI detection problem consists of two sub-tasks: 
1) detecting pairs of interacting human and object (human-object pair detection)
and 2) classification of their interactions.
These two tasks are not independent, but in a cascaded relationship, as shown in Fig.~\ref{fig:problem_describ}.
Specifically, in our toolbox, we first perform a \emph{holistic} analysis of the overall HOI detection accuracy. 
Inspired by the object detection diagnosis toolbox~\cite{bolya2020tide}, we define a set of error types as well as oracles to fix them in the HOI detection pipeline across the human-object pair detection and interaction classification tasks.
The $mAP$ improvement, obtained by applying the oracle to each error, is used to measure the significance of different errors.
The larger $mAP$ improvement can be obtained for a particular type of error, the more it contributes to the failure of an HOI detector.

We then delve into the human-object pair detection and interaction classification tasks, respectively, and conduct detailed studies.
For the detection task, we mainly investigate \emph{Recall} to see if it can detect all the ground-truth human-object pairs for the later stage of interaction classification.
We also compute \emph{Precision} to check the noisiness level of the detections.
For the interaction classification task, an HOI model needs to differentiate negative detections, where the detected human-object pairs have no actual interactions, from positive ones (\ie, with actual interactions).
To diagnose such a binary classification problem, we report the $AP$ (Average Precision) score to avoid selecting a threshold for the classification score, which is non-trivial.
We also compute the $mAP$ scores for the multi-label interaction classification problem, where we assume the human-object pair detections are correct.
In this way, we can disentangle two sub-tasks and focus on analyzing the interaction classification problem only to gain better insights.

Our diagnosis toolbox is applicable to different methods across different datasets. 
Based on both such holistic and detailed investigations of the human-object pair detection and interaction classification, our  toolbox provides a comprehensive diagnosis report for 8 state-of-the-art HOI detection models. 
With the detailed quantitative break-down results, we are now able to answer questions such as ``\emph{Are one-stage HOI detection models superior to two-stage ones or vice versa?}'' (no clear advantage of one paradigm over the other in terms of accuracy), ``\emph{What is the bottleneck of HOI detection?}'' (incorrect localization of the object in a human-object pair and incorrect classification of the interactions), ``\emph{Why does state-of-the-art method RLIPv2~\cite{yuan2023rlipv2} perform better?}'' (since it significantly improves the interaction classification accuracy), etc. Please refer to Section~\ref{sec:discussions} for detailed discussions of existing HOI detection models.

To our best knowledge, this is the first toolbox dedicated for the diagnosis of HOI detection in static images.
We will release our toolbox and believe our work will foster the future development of HOI detection models. %

\subsection{Related Work}
There are several analysis tools for object detection~\cite{lin2014microsoft,hoiem2012diagnosing,bolya2020tide}. The seminal work~\cite{hoiem2012diagnosing} shows how to analyze the influences of object characteristics on detection performance and the impact of different types of false positives. 
But it requires extra annotations to help analyze the impacts of object characteristics, 
which is unlikely to be scalable in large-scale benchmark datasets. 
TIDE~\cite{bolya2020tide} improves the default evaluation tool provided by the COCO dataset~\cite{lin2014microsoft}. It provides a more general framework for quantifying the performance improvement for different false positive and false negative errors in object detection and instance segmentation algorithms. 
Our quantitative analysis of different errors and different tasks in HOI detection is motivated by TIDE~\cite{bolya2020tide}. 
Simply extending such toolboxes to HOI detection is not trivial due to the coupled nature of human-object pair detection and interaction classification sub-tasks.
Moreover, we delve into each of them, examining models' behavior and identifying their bottleneck.

A similar error diagnosis work~\cite{chen2021diagnosing} is proposed for the video relation detection task, which adopts a similar holistic approach inspired by TIDE~\cite{bolya2020tide}. 
In our diagnosis toolbox, we go beyond the holistic error analysis and also conduct detailed investigations in two different sub-tasks of HOI detection, considering the cascade nature of the HOI detection pipeline. 
In \cite{gupta2015visual}, the authors also define several error types of false positives. However, the definition is specifically tailored for the annotation format of the V-COCO dataset, 
which is not generalizable to others. 
In contrast, our analysis is applicable to different benchmark datasets~\cite{chao2018learning,gupta2015visual}.
In~\cite{kilickaya2020diagnosing}, 
the authors analyze a specific issue of HOI detection, the long-tail problem of HOI categories and points out limiting factors. 
\cite{liu2022highlighting} proposes a new metric to advance HOI generalization, preventing the model from learning spurious object-verb correlations.
Both~\cite{kilickaya2020diagnosing} and~\cite{liu2022highlighting} are complementary to our diagnosis tool and analysis results.

\section{Preliminaries}

\newcommand{\calS}{\mathcal{S}}
\subsection{Definition of HOI Detection}
\label{sec:definition}
Given an input image $I$, the output of an HOI detector is a set of triplets $\calS=\left \{ \left ( b_{i}^{h}, b_{i}^{o}, a_{i}\right )\right \}_{i=1}^K$, where $b_{i}^{h}$, $b_{i}^{o}$, and $a_{i}$ denote the $i$-th human bounding box, object bounding box, 
and their interaction class, respectively. 
Both $b_i^h$ and $b_i^o$ contain the coordinates of the bounding boxes as well as the category labels associated with them.
$K$ is the number of triplets. 
In essence, the HOI detection problem consists of two sub-tasks, 
as shown in Fig.~\ref{fig:problem_describ}.
Firstly, it is required to correctly localize every human-object pair that has as any actual interaction.
Unlike object detection, 
the localization task here is to associate a pair of human and object boxes.
We then need to recognize their interaction labels. 
Note that, there may be multiple interactions associated with the same human-object pair, making it a \emph{multi-label classification} problem. 
For instance, the interaction of the person and skateboard in Fig.~\ref{fig:problem_describ} can be \texttt{hold}, \texttt{ride}, etc.

\subsection{Computing $mAP$}
\label{subsec:map}

For an output triplet $\left ( b_{i}^{h}, b_{i}^{o}, a_{i}\right )$ from a model, it is compared with the ground-truth annotations, 
and considered to be a true positive (TP) of a HOI class if \emph{all} the following conditions are satisfied:
\begin{itemize}[noitemsep,topsep=2pt,leftmargin=15pt]%
    \item The category labels of the human and object bounding boxes are both correct.
    \item The intersection-over-union (IoU) w.r.t. the ground-truth annotations for the human $\mathrm{IoU}^{h}$ and object $\mathrm{IoU}^{o}$ both exceed 0.5, {\ie}, $\min \bigl( \mathrm{IoU}^{h}, \mathrm{IoU}^{o} \bigr) > 0.5$.
    \item The output interaction label $a_i$ is correct.
\end{itemize} 
If \emph{any} of them is not satisfied, it is considered as a false positive (FP).
If multiple HOI predictions are matched to the same ground-truth HOI triplet, the one with the highest confidence score, which is defined as the product of confidence (\ie, classification) scores of $b_i^h$, $b_i^o$, and $a_i$, is chosen to be a TP while all others are considered as FPs.

\begin{figure}[t]
    \centering
    \includegraphics[width=0.95\linewidth]{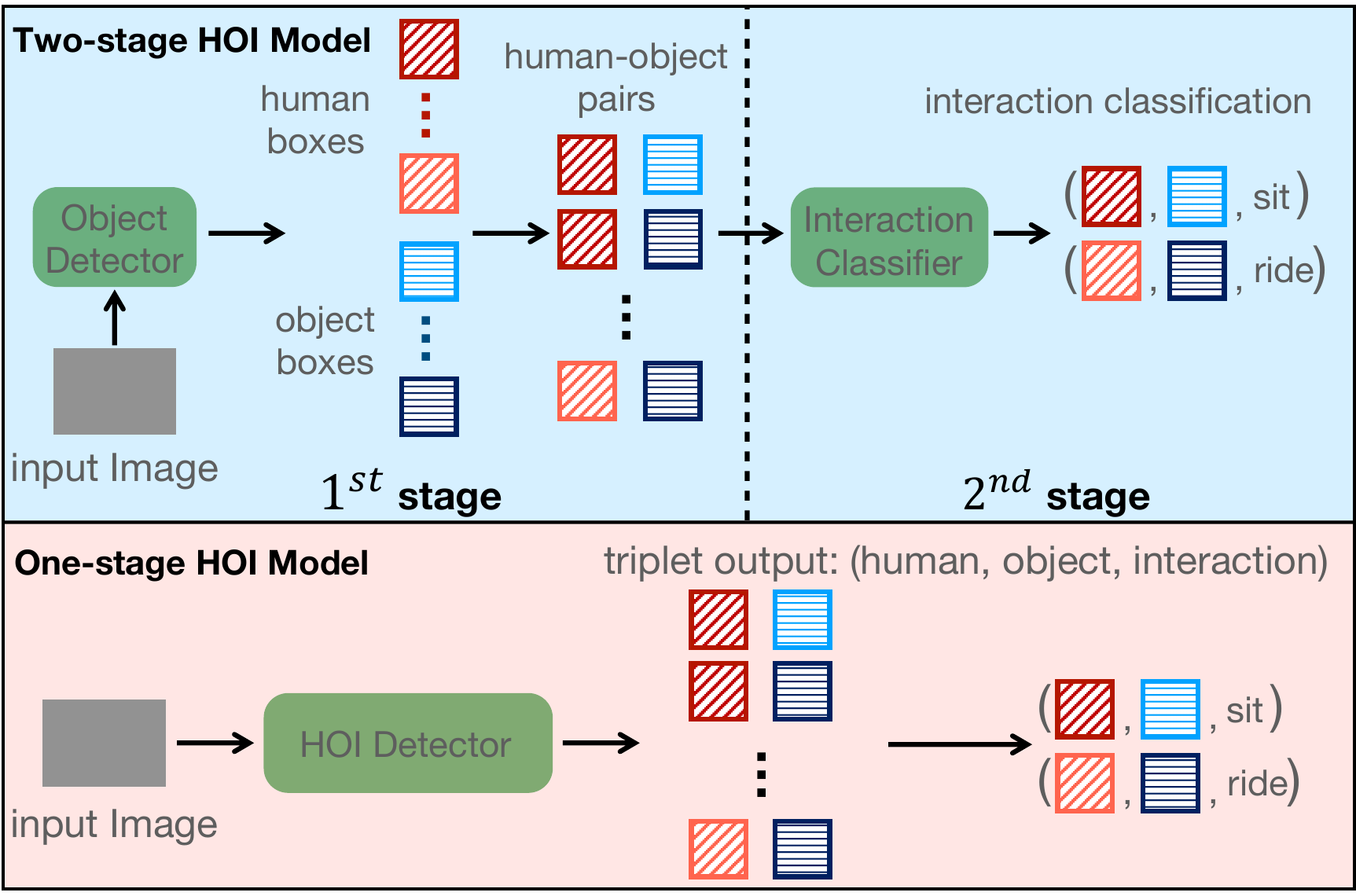}
    \caption{\textbf{Illustration of two- and one-stage HOI detectors.} A two-stage HOI detector separates human-object pair detection and interaction classification, while a one-stage model has no such clear separation and directly outputs detected triplets.}
    \label{fig:one-two-stage-HOI}
\end{figure}

All output triplets are collected from all the images in a benchmark for each HOI category. 
The detected triplets are sorted by descending the confidence scores. 
For each HOI category, given a threshold of the triplet confidence $\tau_t$,
the cumulative precision and recall are defined as %
\begin{equation}
   P=\frac{N_{TP}}{N_{TP}+N_{FP}}, \quad R=\frac{N_{TP}}{{N_{GT}}}, 
  \label{eq:pr}
\end{equation}
for those triplets whose confidence scores are greater than $\tau_t$.
$P$ denotes the precision and $R$ is the recall. 
$N_{TP}$, $N_{FP}$, and $N_{GT}$ are the number of TPs, FPs, and ground-truth triplets in a particular HOI category. By varying the confidence threshold $\tau_t$, $P$ is interpolated such that it decreases monotonically, 
and $AP$ (Average Precision) is computed as the integral under the precision-recall curve. Finally, $mAP$ is defined as the average $AP$ over all HOI categories.

In our diagnosis, we exclude the 80 \nointeraction HOI categories on HICO-DET due to their incomplete annotations.
Even if a model correctly outputs the \nointeraction labels, they will be considered as false positives  and the model’s $mAP$ gets penalized incorrectly.
We discuss this issue in detail in the supplementary.

\subsection{Two-stage \textit{vs.} One-stage HOI Detectors} 
\label{subsec:two_vs_one_stage}
Existing HOI detectors can be roughly grouped into two categories: 
two-stage and one-stage, as illustrated in Fig.~\ref{fig:one-two-stage-HOI}.
Two-stage HOI detectors first detect individual object instances, yielding a set of human and object bounding boxes, whose confidence scores must be greater than a \emph{fixed} threshold $\tau_d$. 
Every single pair of human and object bounding boxes will then be \emph{exhaustively} paired for action classification in the second stage,
where the the object detector's feature representations are used for the interaction classification. 
Depending on the choice of the object detector, NMS (non-maximum suppression) may be adopted to remove duplicate object detections so there are no more duplicates in the human-object pairs and the final triplet outputs.

As for one-stage methods, the human-object pair detection and interaction classification are performed together without clear separation.
A one-stage detector directly localizes the human-object pairs that may have interactions and classifies the interactions between humans and objects, where the feature representations are shared for both tasks. 
NMS is usually adopted to remove duplicates in the final output of detected triplets.

Both two paradigms are being actively investigated. %
One-stage detectors usually run faster than their two-stage counterparts since they skip the individual object detection and couple the human-object pair detection and interaction classification together.
But in terms of accuracy (\ie, $mAP$), there is no clear advantage for one over the other.

\begin{figure*}[t]
    \centering
    \includegraphics[width=\textwidth]{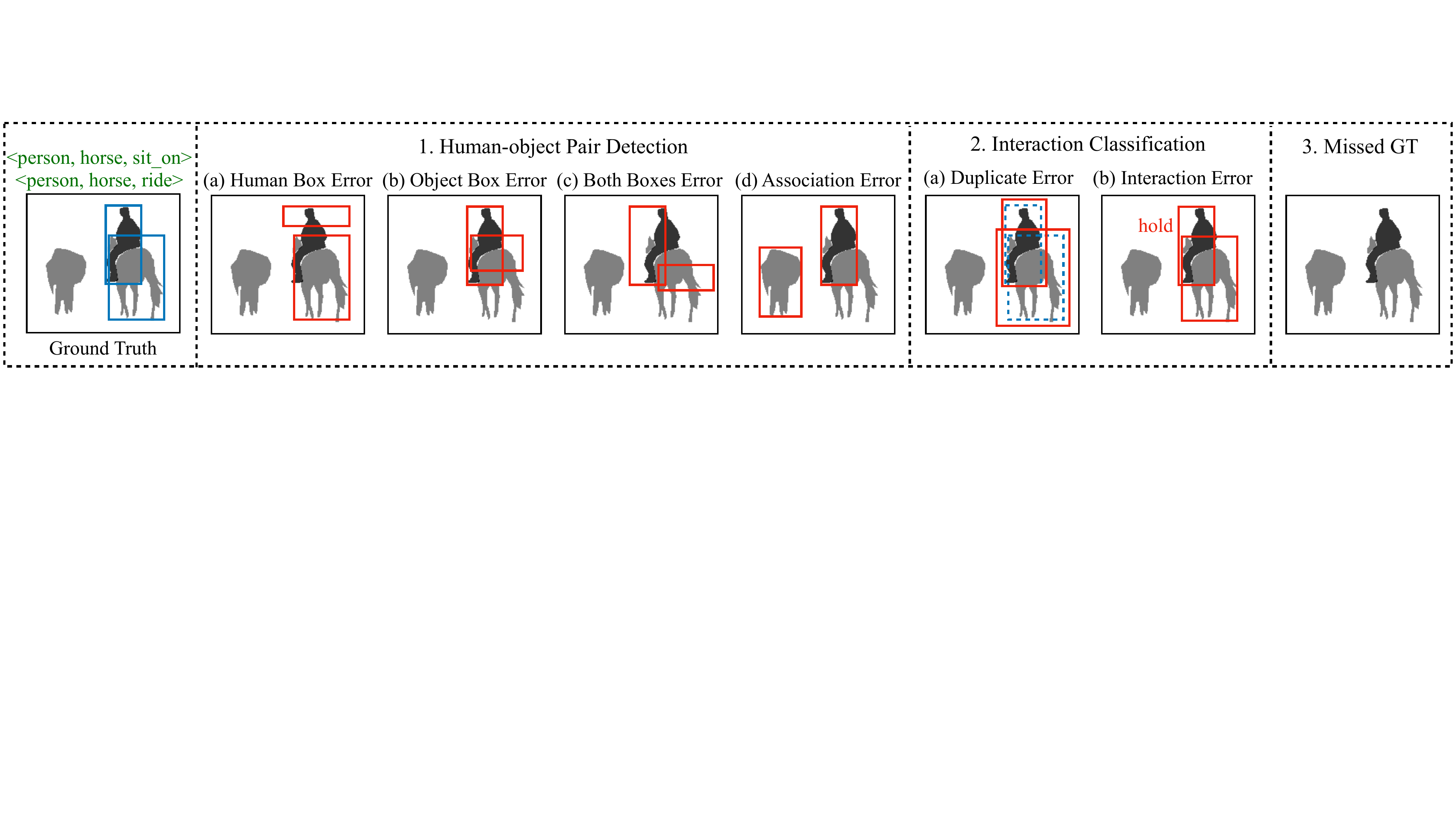}
    \caption{\textbf{Examples of different types of errors.} A human-object pair in the image has two occurring interactions. We show seven sample HOI triplet detections, corresponding to seven types of errors. For clarity, we only show a single triplet detection on each image~(solid line). The dash line means there already exists a detection with a higher confidence score.}
    \label{fig:error_type_vis}
\end{figure*}

\subsection{Benchmark Datasets}
HICO-DET~\cite{chao2018learning} and V-COCO~\cite{gupta2015visual} are two widely used benchmark datasets, both of which share the same 80 object categories as in the COCO dataset~\cite{lin2014microsoft}. HICO-DET has 117 interaction classes,
leading to 600 HOI categories~(some combinations of objects and interactions are not feasible). In V-COCO, there are 26 interaction classes. For each interaction, objects are annotated in three different roles: the agent, the instrument, or the object. 
The task is to detect the agent (human) and the objects in various roles for the interaction (\eg, \texttt{<person cut\_instrument knife>}, 
\texttt{<person read\_object book>}).

\section{Our Diagnosis Toolbox}
\label{sec:method}

\subsection{Holistic Error Analysis}
\label{sec:importance}
One way of diagnosing a method is to investigate the error patterns in its output.
As we can see in Section~\ref{subsec:map}, an HOI detection model can make errors in many places across the human-object pair detection and interaction classification tasks, leading to more FPs and FNs (false negatives) and fewer TPs, and thus lower $mAP$. 
Inspired by the diagnosis method of object detection~\cite{bolya2020tide}, we conduct an holistic analysis by defining a set of error types of HOI detection first as shown in Fig.~\ref{fig:error_type_vis}. 
We then quantify the importance of each error by checking how much $mAP$ improvement could be obtained if such an error were \emph{perfectly} solved using predefined oracles.

\noindent\textbf{Human-object Pair Detection Errors.} 
We define the following set of errors in the human-object pair detection task. 
\begin{itemize}[noitemsep,topsep=2pt,leftmargin=*]%
    \item \textbf{Human box error:} The detected object bounding box is correct, but the human bounding box is incorrect (either incorrect localization where $\mathrm{IoU^h}<0.5$, or incorrect classification of the human label, or both).
    \item \textbf{Object box error. } The detected human bounding box is correct but the object bounding box is incorrect (either incorrect localization where $\mathrm{IoU^o}<0.5$, or incorrect classification of the object label, or both).
    \item \textbf{Both boxes error:} 
    Neither the detected human nor object bounding box is correct.
    \item \textbf{Association error:} 
    Both the human and object bounding boxes are correct, 
    but they have no actual interaction.
\end{itemize}
\noindent\textbf{Interaction Classification Errors.}
Here we only need to consider human-object pairs with actual interactions, which should be correctly localized already (otherwise, such an pair would be captured by an error in the human-object pair detection task). We define the following interaction classification errors.
\begin{itemize}[noitemsep,topsep=2pt,leftmargin=*]%
    \item \textbf{Duplicate error:} The output action label is correct, but there is another detected triplet with a higher confidence score that has already matched to the ground truth.
    \item \textbf{Interaction error:} The output interaction is different from the ground-truth label.
\end{itemize}
\noindent\textbf{Missed GT Error.}
If a ground truth triplet is missing in the HOI detection results and also not covered by any of the above errors, such triplets will be considered as missed GT.

\begin{table*}[t]
\footnotesize
    \centering
    \caption{\label{tab:model-arc}\textbf{Details of HOI detection models used in our analysis}, including four two-stage and four one-stage detectors. They cover a wide range of design choices (\eg, backbone, object/pair detector, and interaction classifier).}
    \begin{tabular}{c|c|c|c|c|c|c|c}
    \toprule
     \multirow{2}{*}{Model} & \multirow{2}{*}{Venue}  & \multirow{2}{*}{Architecture} & \multirow{2}{*}{Backbone} & Object/Pair &
     Interaction & \multicolumn{2}{c}{$mAP$}  \\
    \cline{7-8}
    \cline{7-8}
     & & & & Detector & Classifier & \footnotesize{HICO-DET} & \footnotesize{V-COCO} \\
     \midrule
      SCG~\cite{zhang2021spatially} & ICCV 2021 & \multirow{4}{*}{two-stage} & ResNet-50 & FPN & GNN & 33.3 & 49.4\\
      UPT~\cite{zhang2021efficient} & CVPR 2022 & & ResNet-50/101 & DETR & Transformer & 34.6 / 35.5 & 59.0 / 60.7\\
      STIP~\cite{zhang2022exploring} & CVPR 2022 &  & ResNet-50 & DETR & Transformer & 31.6 & 67.2 \\
      RLIPv2~\cite{yuan2023rlipv2} & ICCV 2023 &  & Swin-T/L & Deform.-DETR & Transformer & 41.6 / 49.1 & 66.3 / 69.5 \\
    \midrule
      CDN~\cite{zhang2021mining} & NeurIPS 2021 & \multirow{4}{*}{one-stage} & ResNet-50/101 & DETR & Transformer & 34.4 / 35.1 & 61.7 / 63.9 \\ 
      GEN-VLKT~\cite{liao2022gen} & CVPR 2022 &  & ResNet-50/101 & DETR & Transformer & 35.8 / 36.5 & 62.4 / 63.6  \\
      QAHOI~\cite{chen2021qahoi} & arXiv 2021 &  & Swin-B/L & Deform.-DETR & Transformer & 35.1 / 37.1 & - \\
      MUREN~\cite{kim2023relational} & CVPR 2023 &  & ResNet-50 & DETR & Transformer & 36.0 & 66.2 \\
    \bottomrule
    \end{tabular}
\end{table*}

\noindent\textbf{Error Significance.} Among all such errors, one may wonder which one is more critical toward improving the $mAP$ of HOI detection. 
To this end, 
we compute the improvement of $mAP$ by fixing each error type with an ``oracle''
\begin{align}
    \Delta mAP_o = mAP_o - mAP,
\end{align}
where $mAP_o$ denotes the $mAP$ score after applying the oracle $o$ to fix an error type (we call it oracle as it is assumed to solve each error perfectly).
Due to the space limit,
we provide details about the oracles for fixing different errors in the appendix.
Here we assume the error is perfectly solved, measuring the most $mAP$ improvement we can get.
The larger the $\Delta mAP_o$ we can see for an error, the more significant it is as a bottleneck for an HOI detection model.

\noindent\textbf{Grouping the Errors.}
In some cases, we may need a more concise summary of the error patterns.
To this end, we group the errors introduced earlier into FPs and FNs, regardless where they stem from in the HOI detection pipeline, and measure the $mAP$ improvement for each of them separately.

\subsection{Diagnosis of Human-object Pair Detection}
For an HOI detector, whether one- or two-stage, 
it relies on the human-object pair detection results for interaction classification.
It is thus important to investigate whether the pair detection results are good enough, \emph{without considering the interaction labels}, disentangling two sub-tasks.

There are two factors impacting the quality of the pair detection: coverage of ground-truth pairs and noiseness level in the detection results.
For the coverage, if a ground-truth human-object pair is missing in the detection results, there is no chance for the interaction classification module to recognize the interaction labels, leading to a FN.
Regarding the noiseness level, if the pair detection results contain too many human-object pairs that have no actual interactions, it will cause significant burden to the interaction module to correctly classify their interaction labels, yielding a lot of FPs.

Specifically, we calculate \emph{Pair Recall} as the percentage of ground-truth human-object pairs that are contained in the detection results. 
Due to the multi-label nature, multiple ground-truth pairs can be matched to the same detected pair. 
In such a case, only one of them is considered into recall computation and other duplicated pairs will be surpressed. 
We finally report the average recall across the entire dataset.

To examine the noiseness level of the detection results, we compute \emph{Pair Precision} as the percentage of detected human-object pairs that are considered as TPs.
Similarly, we report the precision score on the dataset level.

\begin{figure*}[]
\centering
\newcommand{\loadFig}[1]{\includegraphics[height=0.165\linewidth]{#1}}
\renewcommand{\tabcolsep}{0.5pt}
\begin{tabular}{cccc}
\loadFig{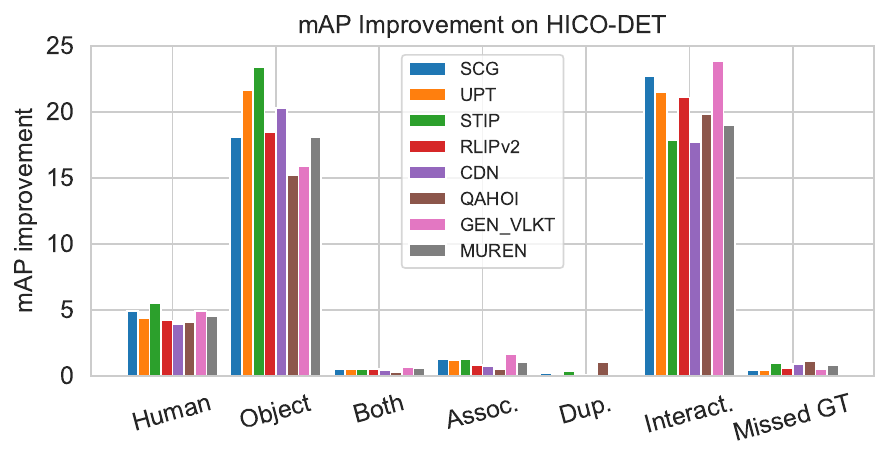} &
\loadFig{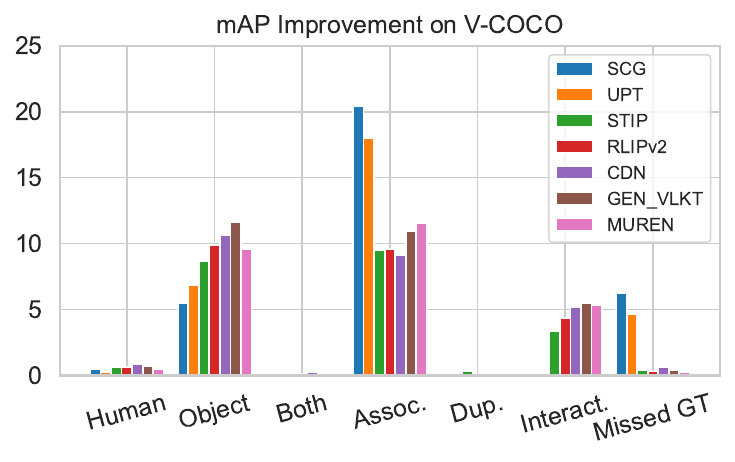} &
\loadFig{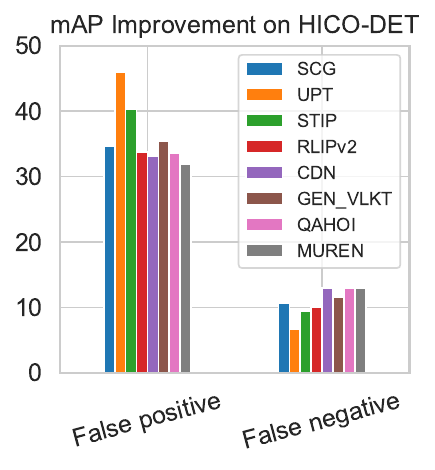} &
\loadFig{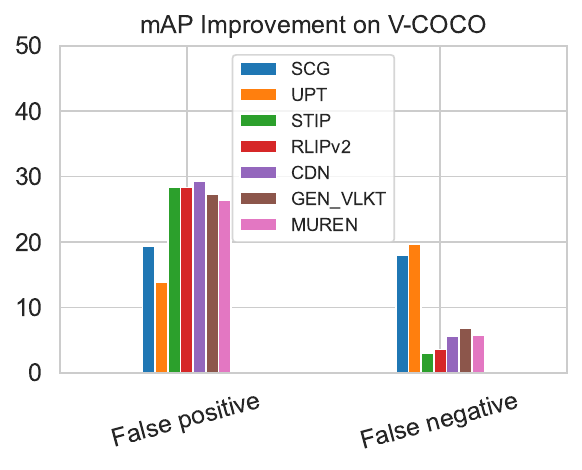} \\
\end{tabular}
\vspace{-10pt}
\caption{\textbf{$mAP$ improvement} by fixing different types of errors on HICO-DET and VCOCO. 
}
\label{fig:mAP_improvement}
\end{figure*}

\begin{figure*}[]
\centering
\newcommand{\loadFig}[1]{\includegraphics[height=0.155\linewidth]{#1}}
\renewcommand{\tabcolsep}{0.5pt}
\begin{tabular}{cccc}
\loadFig{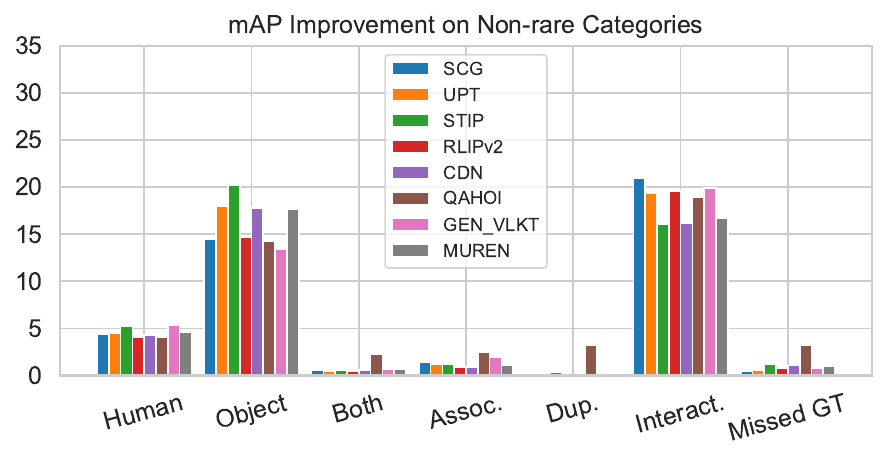} &
\loadFig{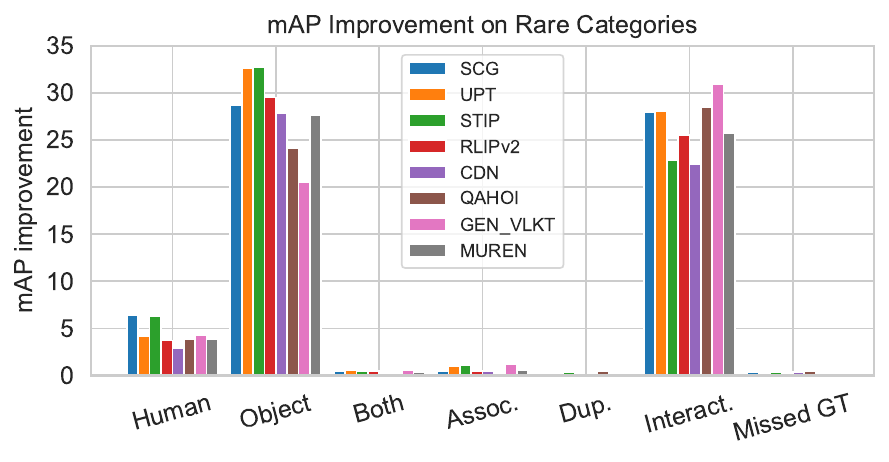} &
\loadFig{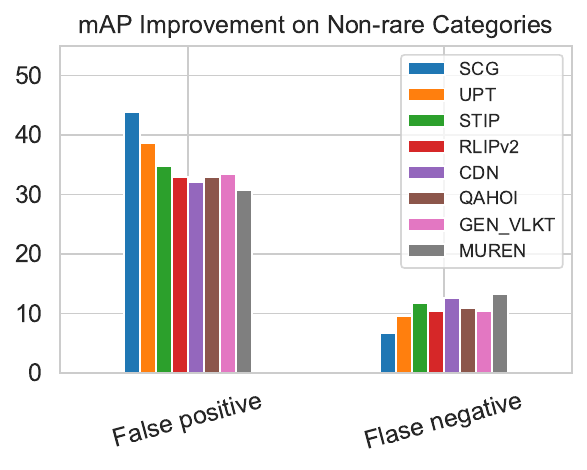} &
\loadFig{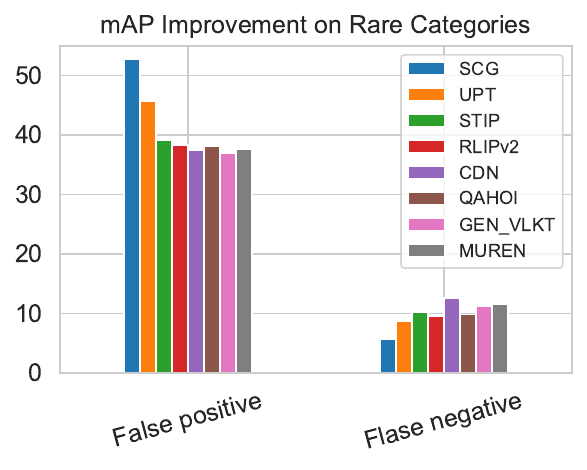} \\
\end{tabular}
\vspace{-10pt}
\caption{\textbf{$mAP$ improvement} by fixing different types of errors for the rare and non-rare HOI categories on HICO-DET. 
}
\label{fig:mAP_improvement_rare_nonrare}
\end{figure*}

\subsection{Diagnosis of Interaction Classification}
\label{subsec:act_classification}

Based on the human-object pair detection results, the interaction classification module needs to handle two cases.

\noindent\textbf{Recognizing incorrect human-object pair detections.}
For those incorrectly localized human-object pairs, they have no actual interactions and should not appear in the final output.
Unlike the multi-label interaction classification, for this task, recognizing an incorrect human-object pair detection is a \emph{binary classification} problem.
In V-COCO, if the classification scores for those actual interaction categories are very low, it indicates the human and object have no actual interactions. 
We therefore compute the classification score belonging to the negative class as $1 - \max_i (p_i)$, where $p_i$ is the classification score of the $i$-th actual interaction class.
We report the $AP$ (average precision) score to avoid the selection of a threshold for the binary classification.

\noindent\textbf{Correct human-object pair detections.}
For correctly localized pairs, there may be multiple interaction labels associated with them, as shown in Fig.~\ref{fig:problem_describ}.
We therefore compute the $mAP$ score for the classification over all possible interaction categories.
Similar to the analysis in the previous section, here we do not consider the detection scores for the correctly detected human-object pairs, \emph{disentangling} the human-object pair detection and interaction classification tasks.

\section{Diagnosis Results}
\subsection{Setup}

In our analysis, 
we diagnose eight popular HOI detection models, include four two-stage and four one-stage, 
covering a wide range of design choices (\eg, backbone, object/pair detector, and interaction classifier). 
We use the code and model weights provided by the authors\footnote{Since the model weights on V-COCO are not available for QAHOI~\cite{chen2021qahoi}, we only report its diagnosis results on HICO-DET.}. 
A summary of these models can be found in Table~\ref{tab:model-arc}.
QPIC~\cite{tamura2021qpic}, HOITR~\cite{zou2021end}, and QAHOI~\cite{chen2021qahoi} share similar model designs based on the DETR object detection model~\cite{carion2020end}.
We therefore only investigate QAHOI here because it reports higher $mAP$.
We provide detailed introductions of these methods in the appendix.

\subsection{Holistic Error Analysis}
The $mAP$ improvement for the seven types of errors as well as FPs and FNs on both HICO-DET and V-COCO are shown in Fig.~\ref{fig:mAP_improvement}.

\definecolor{gain}{HTML}{34a853}  %
\newcommand{\gain}[1]{\textcolor{gain}{#1}}
\definecolor{lost}{HTML}{ea4335}  %
\newcommand{\lost}[1]{\textcolor{lost}{#1}}
\newcommand{\res}[2]{{#1} {({\gain{#2}})}}
\newcommand{\bad}[2]{{#1} {({\lost{--#2}})}}

\begin{table*}[t]
\centering
\caption{\label{tab:recall_prec_hico_vcoco}
\textbf{Diagnosis results of human-object pair detection and interaction classification.}
``Neg. $AP$'' captures the performance of classifying the negative human-object pairs that have no actual interactions. ``Inter. $mAP$'' is about the interaction classification for the correctly detected pairs. ``HOI $mAP$'' is about the entire HOI detection task.
 } 
\renewcommand{\tabcolsep}{6pt}
\newcolumntype{?}{!{\vrule width 1pt}}
\resizebox{0.99\textwidth}{!}{

\begin{tabular}{c|c|c?c|c|c|c|c?c|c|c|c|c|c|c}
\toprule
\multirow{2}{*}{Arch.} & \multirow{2}{*}{Method} & \multirow{2}{*}{Backbone} & \multicolumn{6}{c|}{HICO-DET } & \multicolumn{6}{c}{V-COCO} \\
\cmidrule{4-9}\cmidrule{10-15}
& & & \#Pairs & Pair Rec. & Pair Prec. & Neg. $AP$ & Inter. $mAP$ & HOI $mAP$ & \#Pairs & Pair Rec. & Pair Prec. & Neg $AP$ & Inter. $mAP$ & HOI $mAP$ \\
\midrule
\multirow{6}{*}{two-stage} & SCG~\cite{zhang2021spatially} & ResNet50 & 44 & 84.0 & 3.7 & 91.8 & 46.5 & 33.3 & 406 & 91.2 & 0.6 & 99.0 & 63.5 & 49.4\\
 & UPT~\cite{zhang2021efficient} & ResNet50 & 46 & 80.1 & 3.7 & 91.9 & 32.5 & 34.6 & 24 & 92.8 & 10.4 & 99.2 & 64.5 & 59.0 \\
 & UPT~\cite{zhang2021efficient} & ResNet101 & 43 & 80.8 & 3.6 & 91.8 & 32.4 & 35.5 & 26 & 93.2 & 7.8 & 99.4 & 64.7 & 60.7 \\
 & STIP~\cite{zhang2022exploring} & ResNet50 & 32 & 74.2 & 4.4 & 91.7 & 51.1 & 31.6 & 32 & 94 & 7.4 & 97.8 & 82.1 & 67.2 \\
 & RLIPv2~\cite{yuan2023rlipv2} & SwinT & 44 & 80.8 & 3.4 & 74.6 & 56.7 & 41.6 & 100 & 93.1 & 2.3 & 96.9 & 78.5 & 66.3 \\
 & RLIPv2~\cite{yuan2023rlipv2} & SwinL & 45 & 84.0 & 3.5 & 74.4 & 61.1 & 49.1 & 100 & 94.0 & 2.4 & 96.8 & 81.5 & 69.5 \\
\midrule
 \multirow{5}{*}{one-stage} & CDN~\cite{zhang2021mining} & ResNet50 & 64 & 79.4 & 2.5 & 67.8 & 54.3 & 34.4 & 100 & 89.7 & 2.3 & 96.8 & 76.6 & 61.7 \\
 & GEN-VLKT~\cite{liao2022gen} & ResNet50 & 64 & 81.1 & 2.6 & 67.3 & 49.6 & 35.8 & 64 & 88.5 & 3.5 & 79.2 & 75.5 & 62.4 \\
 & GEN-VLKT~\cite{liao2022gen} & ResNet101 & 64 & 81.8 & 2.6 & 70.0 & 52.5 & 36.5 & 64 & 89.5 & 3.4 & 86.7 & 77.2 & 63.6 \\
 & QAHOI~\cite{chen2021qahoi} & SwinB & 64 & 81.6 & 1.6 & 84.1 & 53.1 & 35.1 & - & - & - & - & - & -\\
  & MUREN~\cite{kim2023relational} & ResNet50 & 64 & 80.1 & 2.5 & 69.8 & 54.5 & 36.0 & 100 & 94.0 & 2.4 & 95.4 & 78.7 & 66.2\\
\bottomrule
\end{tabular}

}
\end{table*}

On HICO-DET, among all the seven errors across the human-object pair detection and interaction classification, we can find that two signifcant errors are shared by all the HOI detectors, no matter one-stage or two-stage: object box error and incorrect interaction classification error.
These can be attributed to two main factors. 
First, HICO-DET has no overlap with the datasets, such as COCO, which are usually used to pre-train the object detector or backbone.
The models therefore tend to incorrectly localize objects in human-object pairs.
Second, HICO-DET comprises a considerable number of interactions, with a significant portion being multi-labeled, posing a challenge for the models to accurately distinguish them.

On V-COCO, the errors are primarily concentrated on the object box errors and association errors of human and object bounding boxes. 
Particularly, two-stage HOI detectors SCG and UPT exhibit a higher frequency of association errors compared to others. %
And overall the $mAP$ improvement on V-COCO is not as notable as on HICO-DET. 
Part of the reason is that detectors/backbones are often pre-trained on the COCO dataset, which V-COCO is built upon. 
As a result, how to correctly associate the detections is more prominent than detecting them.
Moreover, there are a relatively smaller and simpler set of interactions in V-COCO compared with the more abstract ones in HICO-DET (\eg, \texttt{inspect, wield}). 
It is worth noting that SCG~\cite{zhang2021spatially} and UPT~\cite{zhang2021efficient} do not exhibit interaction errors on V-COCO due to their pre-processing and suppression techniques.

In terms of FPs and FNs, from the last two figures of Fig.~\ref{fig:mAP_improvement}, we can see that for most HOI detectors on both HICO-DET and V-COCO, suppressing FPs brings significantly higher $mAP$ improvement than FNs, except for SCG~\cite{zhang2021spatially} and UPT~\cite{zhang2021efficient} on V-COCO. 
It suggests that the existence of incorrect tripelts in HOI detection results is holding the existing models back more than missing the ground-truth triplets.

\subsection{Human-object Pair Detection}
We report the average Pair Recall and Pair Precision for the human-object pair detection task on both HICO-DET and V-COCO in Tab.~\ref{tab:recall_prec_hico_vcoco}.

Perhaps a little surprisingly, two-stage models produce fewer human-object pairs than the one-stage counterparts, even though they exhaustively pair all detected human and object bounding boxes. 
On both HICO-DET and V-COCO, two-stage models tend to have higher Pair Precision scores, indicating less noises (incorrect human-object pairs) in the detection results.
This is partially because in two-stage models, the NMS is usually applied before the pairing of human and object bounding boxes, as introduced in Sec.~\ref{subsec:two_vs_one_stage}, which removes duplicates in the human-object pair detection results.
To examine this factor, we apply NMS to remove the duplicate human-object pair detections for one-stage models, too.
Due to limited space, the results are put in the appendix. 
After doing so, we indeed see fewer human-object pairs detected, decreased Pair Recall, and increased Pair Precision.
We also examine increasing the number of human-object pairs for two-stage models by lowering the object detection threshold $\tau_d$.
As expected, we see reverse effect compared to one-stage models, where we can see more human-object pairs, increased Pair Recall, and decreased Pair Precision.
However, such changes in the human-object pair detection part does not lead to significant change of the final HOI detection $mAP$, indicating the bottleneck in the subsequent interaction classification part, which we will analyze in the following section.

It is worth noting that the Pair Recall values for both one-stage and two-stage methods are notably lower than 100, indicating a lot of human-object pairs are discarded in this stage that have no chance to appear in the final output.

\subsection{Interaction Classification}
In Tab.~\ref{tab:recall_prec_hico_vcoco}, we report $AP$ of classifying negative human-object pairs (Neg. $AP$) to see if the model is able to suppress incorrectly detected human-object pairs by giving them low confidence scores. 
As we can see, two-stage methods perform better on this task on both HICO-DET and V-COCO. 
From the results of interaction $mAP$ (Inter. $mAP$) in Tab.~\ref{tab:recall_prec_hico_vcoco}, we can see that although two-stage models have relatively higher Pair Precision, their interaction classification heads are hard to correctly classify all the interactions (except for RLIPv2).
Instead, one-stage models output more confident scores towards correct interaction predictions, leading to higher Inter. $mAP$. 
The advantages of two-stage \emph{vs.} one-stage models in human-object pair detection and interaction classification get canceled. As a result, the overall HOI detection $mAP$ for both two-stage and one-stage models are roughly the same (except for RLIPv2).

RLIPv2 achieves significantly higher HOI detection $mAP$ than other methods, including both two-stage and one-stage ones.
From Tab.~\ref{tab:recall_prec_hico_vcoco}, we can see that its main advantage is about the substantially higher Inter. $mAP$, which is mainly because its usage of large-scale relational data (VG~\cite{krishna2017visual}, COCO~\cite{lin2014microsoft}, and Object365~\cite{shao2019object365}) for pre-training.

\section{Discussions}
\label{sec:discussions}
\noindent\textbf{Two-stage vs. one-stage HOI detection models.}
In terms of overall HOI detection $mAP$, there is no clear advantage of one paradigm over the other.
With our diagnosis toolbox, we can derive more insights about such two types of HOI detectors.
On the one hand, according to our holistic error analysis, the $mAP$ improvement 
shown in Fig.~\ref{fig:mAP_improvement} demonstrate that both two-stage and one-stage detectors share the same bottleneck in the pipeline, where they both struggle with significant errors in detecting the object in a human-object pair and getting accurate interaction classification when the pair detection is correct. 
And FPs are more dominent than FNs in the errors.

On the other hand, interestingly, after delving into the human-object detection and interaction classification tasks, we can see two paradigms have their own advantages.
As reported in Tab.~\ref{tab:recall_prec_hico_vcoco}, two-stage models generally detect human-object pairs with similar Pair Recall but higher Pair Precision values than one-stage counterparts, indicating the detection results are less noisy.
At the same time, while two-stage models are better at recognizing negative human-object pairs without actual interactions (Neg. $AP$), one-stage methods are superior in recognizing the actual interactions for those correctly localized human-object pairs (Inter. $mAP$).
As a result, the final HOI detection $mAP$ scores are roughly the same for two types of HOI detection models in general (except for RLIPV2, which will be discussed later).

\noindent\textbf{Different backbones.} A stronger backbone can help improve the $mAP$ score of HOI detection. But where is the improvement from?
To answer this questions, we study three methods, UPT~\cite{zhang2021efficient} (ResNet50 \emph{vs.} ResNet101), RLIPv2~\cite{yuan2023rlipv2} (SwinT \emph{vs.} SwinL), and GEN-VLKT~\cite{liao2022gen} (ResNet50 \emph{vs.} ResNet101).
According to Tab.~\ref{tab:recall_prec_hico_vcoco}, a better backbone for UPT mainly leads to slightly better Pair Recall in terms of human-object pair detection, which improves the final HOI $mAP$ on both HICO-DET and V-COCO.
Whereas for RLIPv2 and GEN-VLKT, the improvement are across the entire HOI detection pipeline: better Pair Recall and interaction classification (Inter. $mAP$).
Moreover, the better backbone also enhances the discrimination capability of discarding incorrectly detected human-object pairs for GEN-VLKT (better Neg. $AP$).

\begin{table*}[t]
\centering
\caption{\label{tab:rare_non_rare}
\textbf{Diagnosis results for rare and non-are HOI categories on HICO-DET.}
} 
\renewcommand{\tabcolsep}{4pt}
\newcolumntype{?}{!{\vrule width 1pt}}
\resizebox{0.90\textwidth}{!}{
\begin{tabular}{c|c|c|c|c|c|c|c|c|c|c}
\toprule
\multirow{2}{*}{Arch.} & \multirow{2}{*}{Method} & \multirow{2}{*}{Backbone} & \multicolumn{4}{c|}{Non-rare Categories} & \multicolumn{4}{c}{Rare Categories} \\
\cmidrule{4-7}\cmidrule{8-11}
& & &  Pair Rec. & Pair Prec. & Inter. $mAP$ & HOI $mAP$ &  Pair Rec. & Pair Prec.  & Inter. $mAP$ & HOI $mAP$\\
\midrule
\multirow{6}{*}{two-stage} & SCG~\cite{zhang2021spatially} & ResNet50 & 84.3 & 3.60 & 41.4 & 36.1 & 72.9 & 0.16 &  13.0 & 25.5\\
& UPT~\cite{zhang2021efficient} & ResNet50 & 78.6  & 1.12 & 26.3 & 36.7 & 70.6  & 0.06 & 7.5 & 27.6\\
& UPT~\cite{zhang2021efficient} & ResNet101 & 80.0 & 1.21 &  26.7 &  36.9& 72.3 & 0.07 & 7.5 & 30.0\\
 & STIP~\cite{zhang2022exploring} & ResNet50 & 73.7 & 1.50 & 40.1  & 32.9 & 66.8 & 0.08 & 12.6 & 29.0\\
 & RLIPv2~\cite{yuan2023rlipv2} & SwinT & 83.4 & 0.85 & 49.0 & 44.1 & 75.6 & 0.05 & 14.6 & 35.0\\
 & RLIPv2~\cite{yuan2023rlipv2} & SwinL & 87.3 & 0.89 &  54.4 & 50.4& 81.2 & 0.05 & 21.7 & 44.8\\
\midrule
\multirow{5}{*}{one-stage} & CDN~\cite{zhang2021mining} & ResNet50 & 77.4 & 0.79 & 44.5 & 36.5 & 72.3 & 0.04 & 12.8 & 29.4\\
 & GEN-VLKT~\cite{liao2022gen} & ResNet50 & 79.5 & 0.81 & 43.1 & 38.8 & 71.7  & 0.04 & 19.5 & 33.1\\
 & GEN-VLKT~\cite{liao2022gen} & ResNet101 & 80.5 & 0.83 & 45.6  & 40.2 & 74.2 & 0.05 & 22.4 & 33.8\\
  & QAHOI~\cite{chen2021qahoi} & Swin-B & 81.0 & 0.53 & 45.0 & 38.4 &  70.4 & 0.03 & 10.5 & 27.1\\
 & MUREN~\cite{kim2023relational} & ResNet50 & 78.6 & 0.80 & 44.4 & 38.2 & 71.5 & 0.04 & 13.7 & 30.7\\
\bottomrule
\end{tabular}}
\end{table*}

\begin{figure}[t]
\centering
\newcommand{\loadFig}[1]{\includegraphics[width=0.75\linewidth]{#1}}
\renewcommand{\tabcolsep}{0.5pt}
\vspace{-4mm}
\begin{tabular}{c}
\loadFig{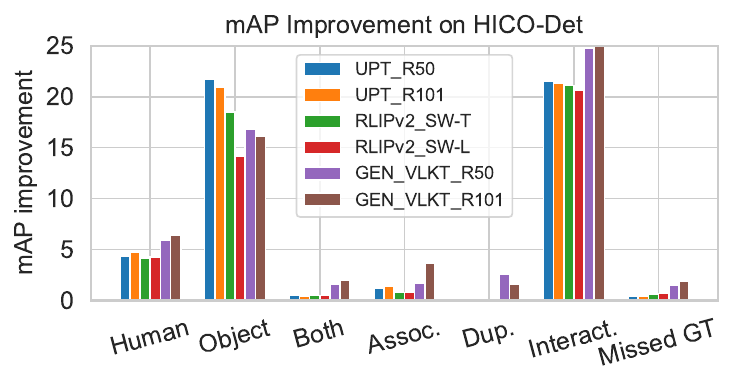} \\
\vspace{-8pt}
\loadFig{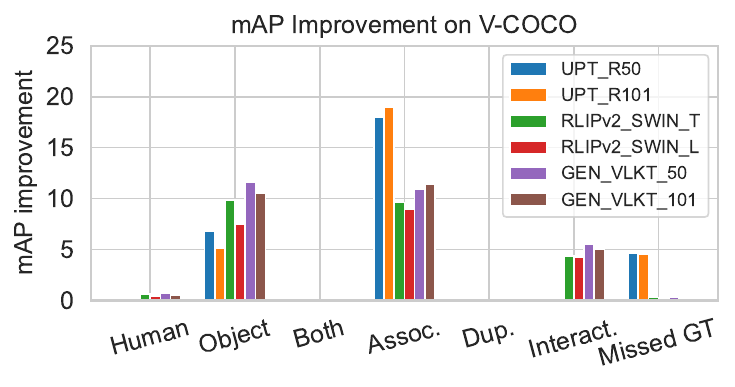} \\
\end{tabular}
\vspace{-10pt}
\caption{\textbf{$mAP$ improvement} of different backbones on HICO-DET and V-COCO. 
}
\label{fig:mAP_improvement_diff_backbone}
\vspace{-12pt}
\end{figure}

In addition, as shown in Fig.~\ref{fig:mAP_improvement_diff_backbone}, our holistic error analysis shows that 
for UPT, RLIPV2, and GEN-VLKT, better backbones lead to less error significance (due to less $mAP$ improvement) for incorrect object localization in a human-object pair on both HICO-DET and V-COCO.
Particularly, for the state-of-the-art RLIPv2 model, switching to a better backbone also leads to less error significance for incorrect interaction classification on both HICO-DET and V-COCO and association error on V-COCO.
However, switching to stronger backbones may not always help for all the errors. 
For instance, slightly more error significance can be seen on incorrect human detection for UPT and interaction classification for GEN-VLKT on HICO-DET.
This explains why the overall HOI detection $mAP$ improvement by the better bones for such two methods are not that significant (0.9 and 0.7, respectively, \emph{vs.} 7.5 of RLIPv2 on HICO-DET).

\noindent\textbf{Rare \textit{vs.} Non-Rare HOI Categories.} The HOI categories follow a long-tail distribution, where some interaction and object classes (\eg, \texttt{ride horse}) are more frequent than others (\eg, \texttt{chase chat}).
On HICO-DET, the HOI categories are divided into rare and non-rare ones, where an HOI category is considered as rare if it has less ten training instances. 
How does the abundance of training instances affect the models' performance?
According to our holistic error analysis in Fig.~\ref{fig:mAP_improvement_rare_nonrare}, the overall distribution of error significance are the same on both rare and non-rare HOI categories.
For instance, incorrect object detection in a human-object pair and incorrect interaction classification are the main bottleneck and FPs are more prominent than FNs.
But the HOI models tend to have more failures in rare HOI categories due to fewer training instances and fixing them leads to larger $mAP$ improvement. 

Furthermore, according to Tab.~\ref{tab:rare_non_rare}, because of less training data are available for rare HOI categories, the accuracy of human-object pair detection (Pair Recall and Pair Precision) and interaction classification (Inter. $mAP$ 
) are consistently lower on rare categories.
Even for the state-of-the-art model RLIPv2~\cite{yuan2023rlipv2} with a strong SwinL backbone, the interaction classification accuracy (Inter. $mAP$) significantly decreases (54.4 \emph{vs.} 21.7).

Both object categories and interaction labels follow long-tail distributions, 
Their combination makes the long-tail issue more severe.
Improving the accuracy on the rare HOI categories (\ie, tail classes) is an open problem~\cite{kilickaya2020diagnosing,liao2022gen,hou2021detecting}.

\noindent\textbf{Performance on HICO-DET \emph{vs.} V-COCO}. Our holistic error analysis in Fig.~\ref{fig:mAP_improvement} shows that the major error significance on both of them are similar.
Our diagnosis in Tab.~\ref{tab:recall_prec_hico_vcoco} further reveals that existing methods tend to produce more human-object pairs to cover the ground truths (higher Pair Recall) on V-COCO than HICO-DET although the noiseness level are roughly the same (Pair Precision).
At the same time, existing methods tend to perform better on the interaction classification task better on V-COCO than HICO-DET, in terms of both discarding incorrectly detected human-object pairs (Neg. $AP$) and recognizing the actual interactions of the correctly detected pairs (Inter. $mAP$).
As we discussed earlier, part of the reason is because of the overlap between V-COCO and COCO, which is usually adopted for detector/backbone pre-training.

\noindent\textbf{Human-object pair detection \emph{vs.} interaction classification.}
Both of our holistic error analysis via $mAP$ improvement and breakdown inspection of human-object pair detection and interaction classification show that there are significant bottleneck in each of the two sub-tasks.
For instance, in Tab.~\ref{tab:recall_prec_hico_vcoco}, we can see that the Pair Recall on HICO-DET are still much lower than 100, implying a lot of ground-truth pairs can not be detected. 
The low Inter. $mAP$ also suggests that correctly recognizing the actual interactions is also a challenging problem.

Practically, human-object pair detection largely depends on the progress of generic single-object detection, which has attracted a significant amount of research attention.
The performance on standard benchmarks (\eg, COCO) is almost saturated. 
In contrast, the interaction classification is a multi-label classification problem, which has not yet been extensively studied.
The success of the state-of-the-art RLIPv2 model~\cite{yuan2023rlipv2} shows great potential to solve this problem by using large-scale relational data for pre-training.

\section{Conclusion}

In this paper, we introduced the first diagnosis toolbox for HOI detectors. 
We first conduct a holistic error analysis by defining a set of errors across the pipeline of HOI detection and report the $mAP$ improvement by fixing each of them using an oracle.
We then delve into the human-object pair detection and interaction classification tasks separately, and provide detailed breakdown inspection for each of them.
Detailed analyses are reported on both HICO-DET and V-COCO over eight state-of-the-art HOI detectors.
We believe our diagnosis toolbox and analysis results will be helpful for fostering future research in this direction.
\clearpage

\appendix

\appendix

\begin{figure}[t]
\centering
\renewcommand{\tabcolsep}{1pt}
\newcommand{\loadFig}[1]{\includegraphics[width=0.45\linewidth]{#1}}
\begin{tabular}{cc}
\loadFig{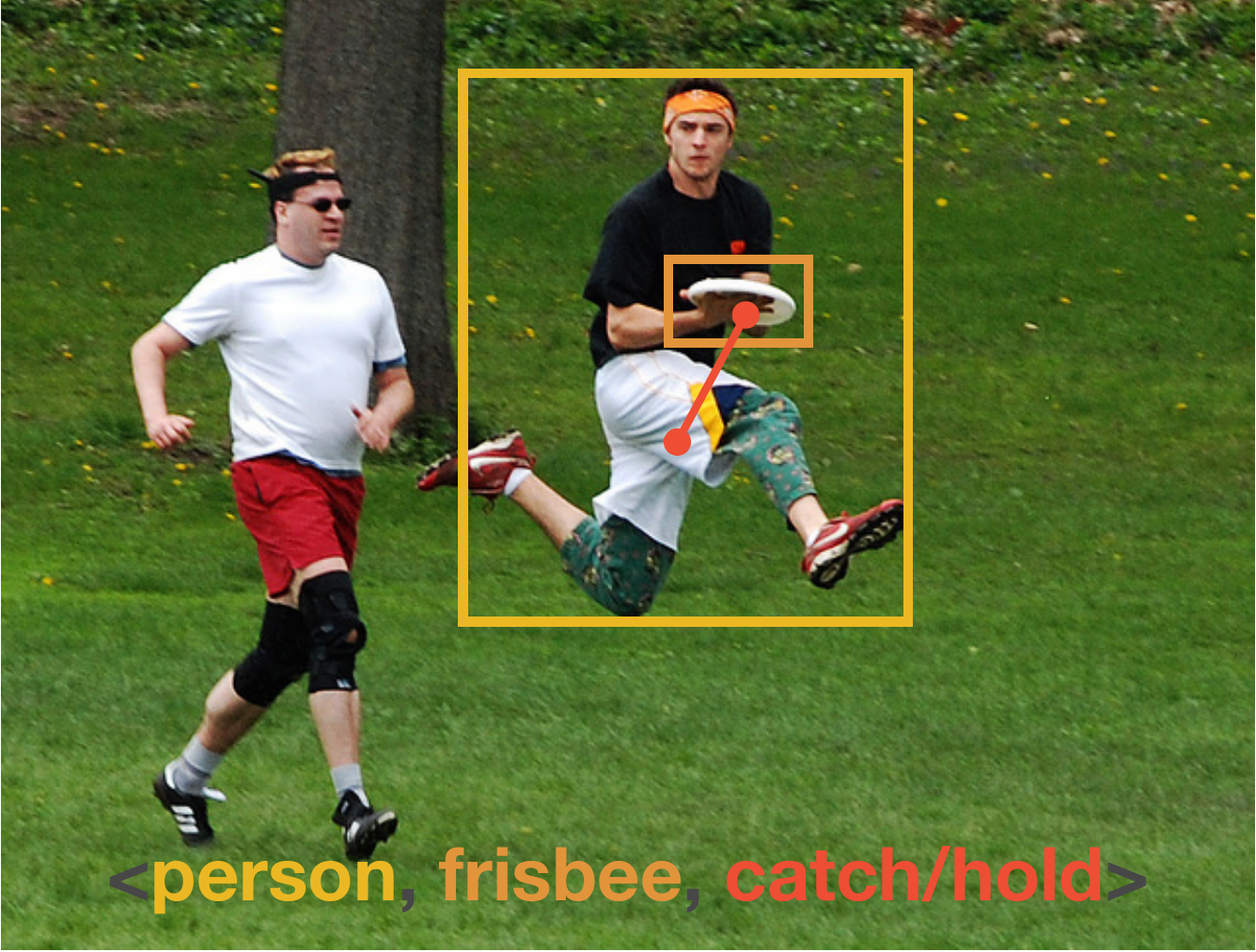} &
\loadFig{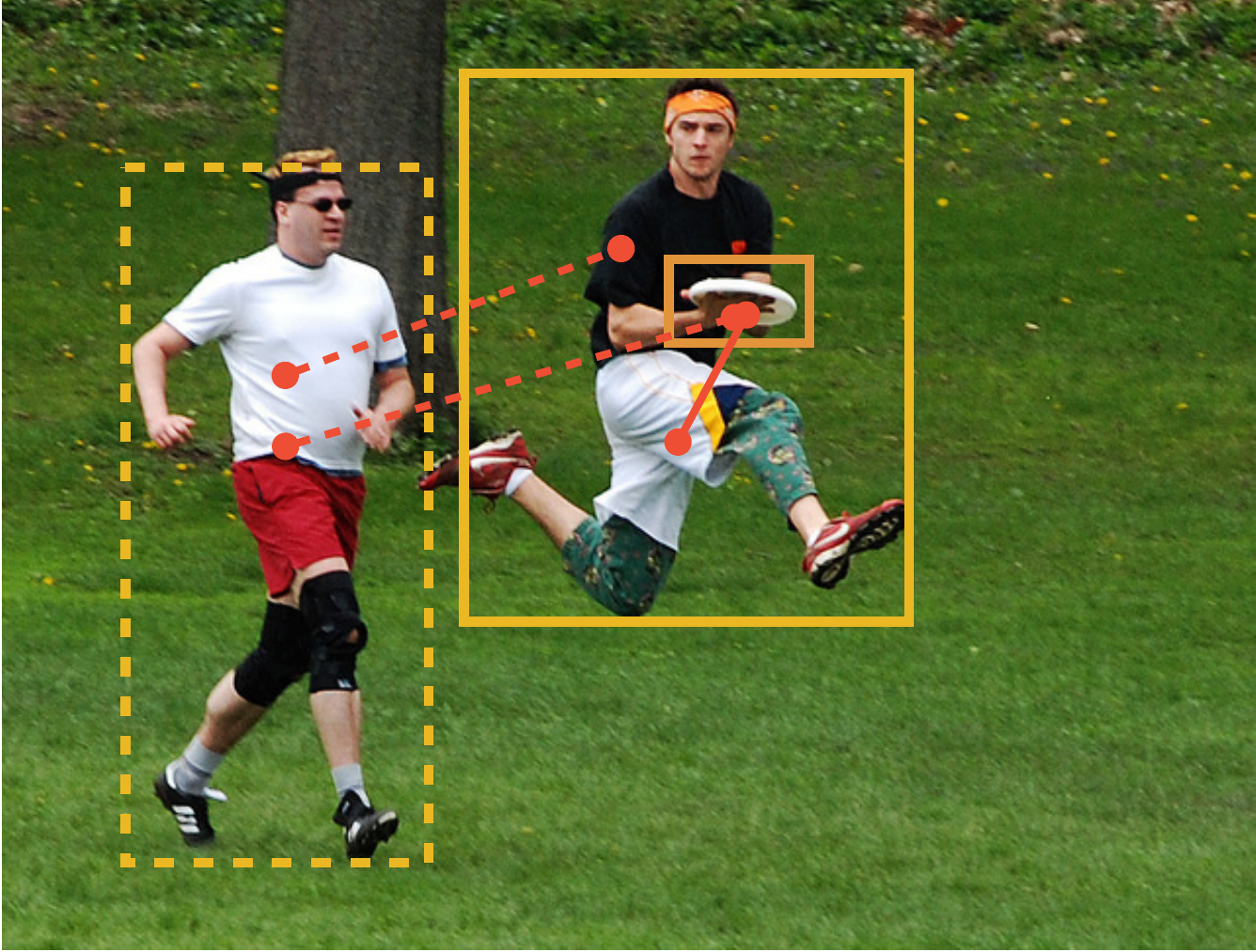} \\
\loadFig{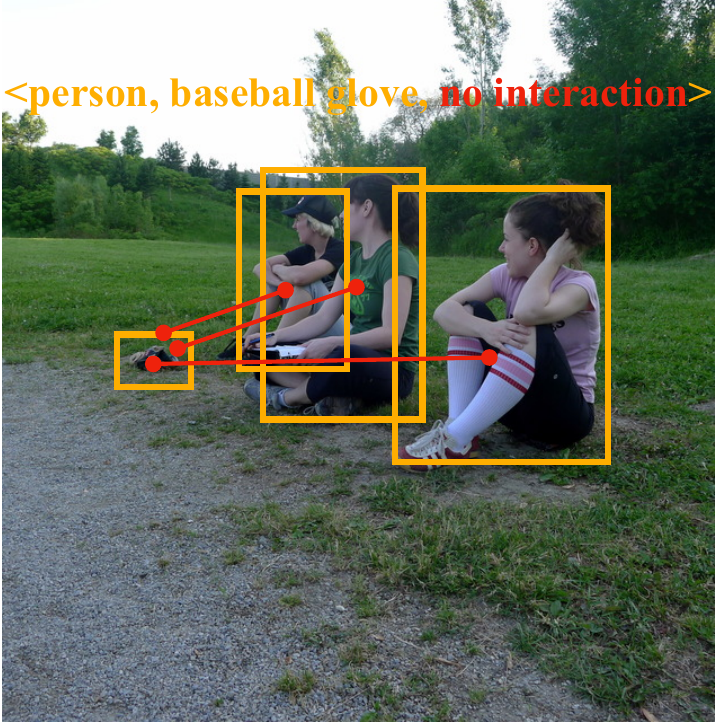} &
\loadFig{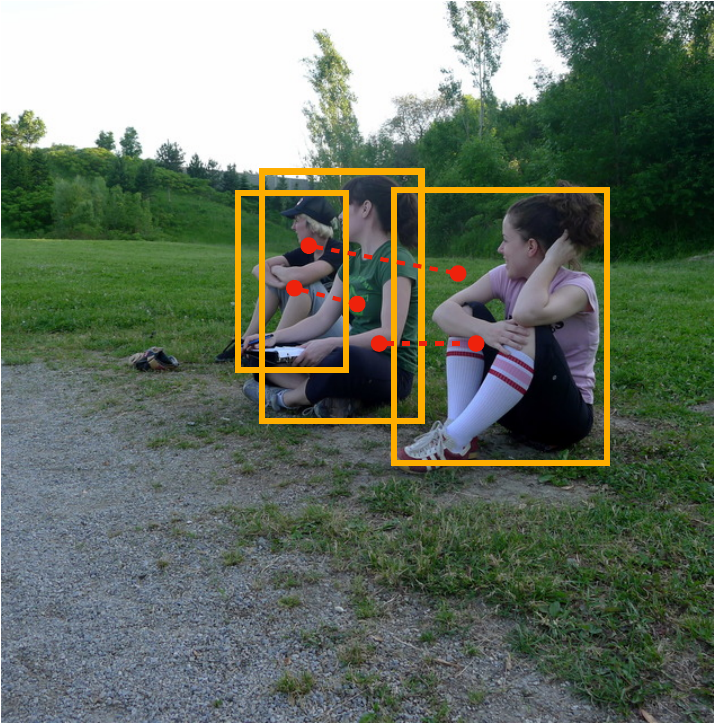} \\
\loadFig{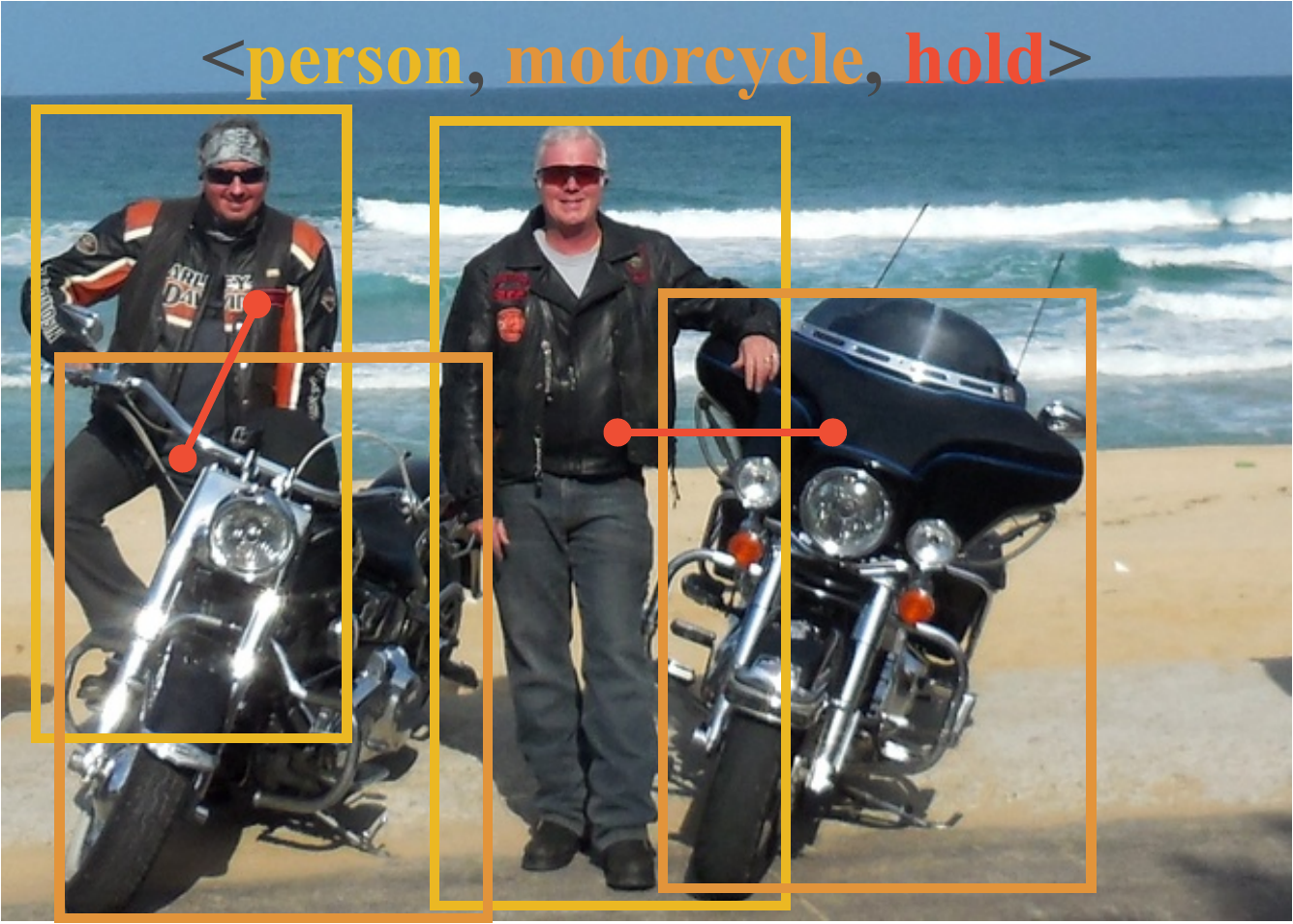} &
\loadFig{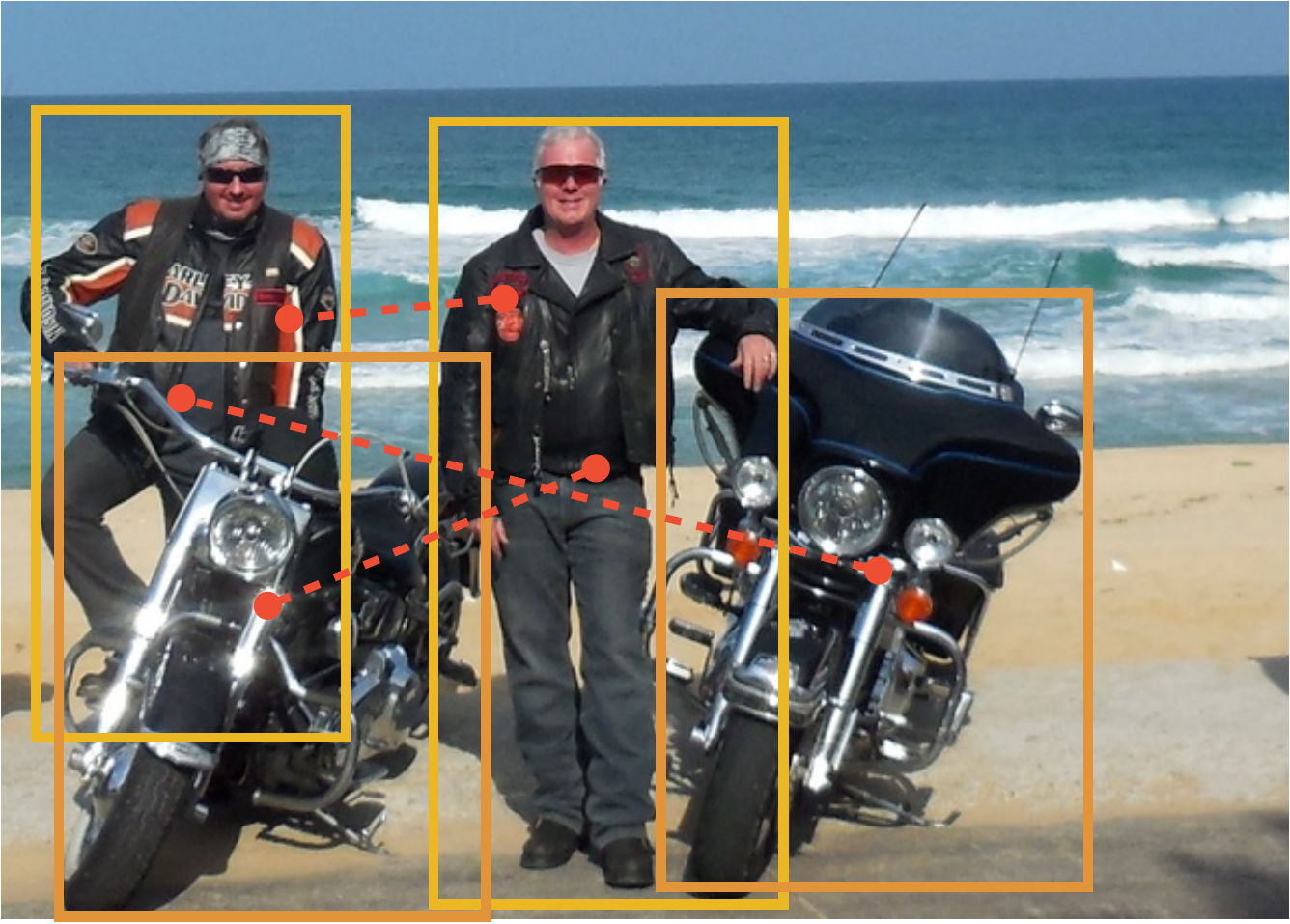} \\
(a) current annotations &
(b) missing annotations \\

\end{tabular}
\vspace{-2pt}
\caption{\textbf{Examples of the missing annotations of the \nointeraction HOI class.} 
On the right, we show missing \nointeraction labels and missing bounding box using dashed lines and bounding boxes, respectively. }
\label{fig:no_interaction_example}
\end{figure}

\section{Elaboration on \nointeraction~Class in HICO-DET}

\begin{figure*}[!htb]
    \centering
    \includegraphics[width=\linewidth]{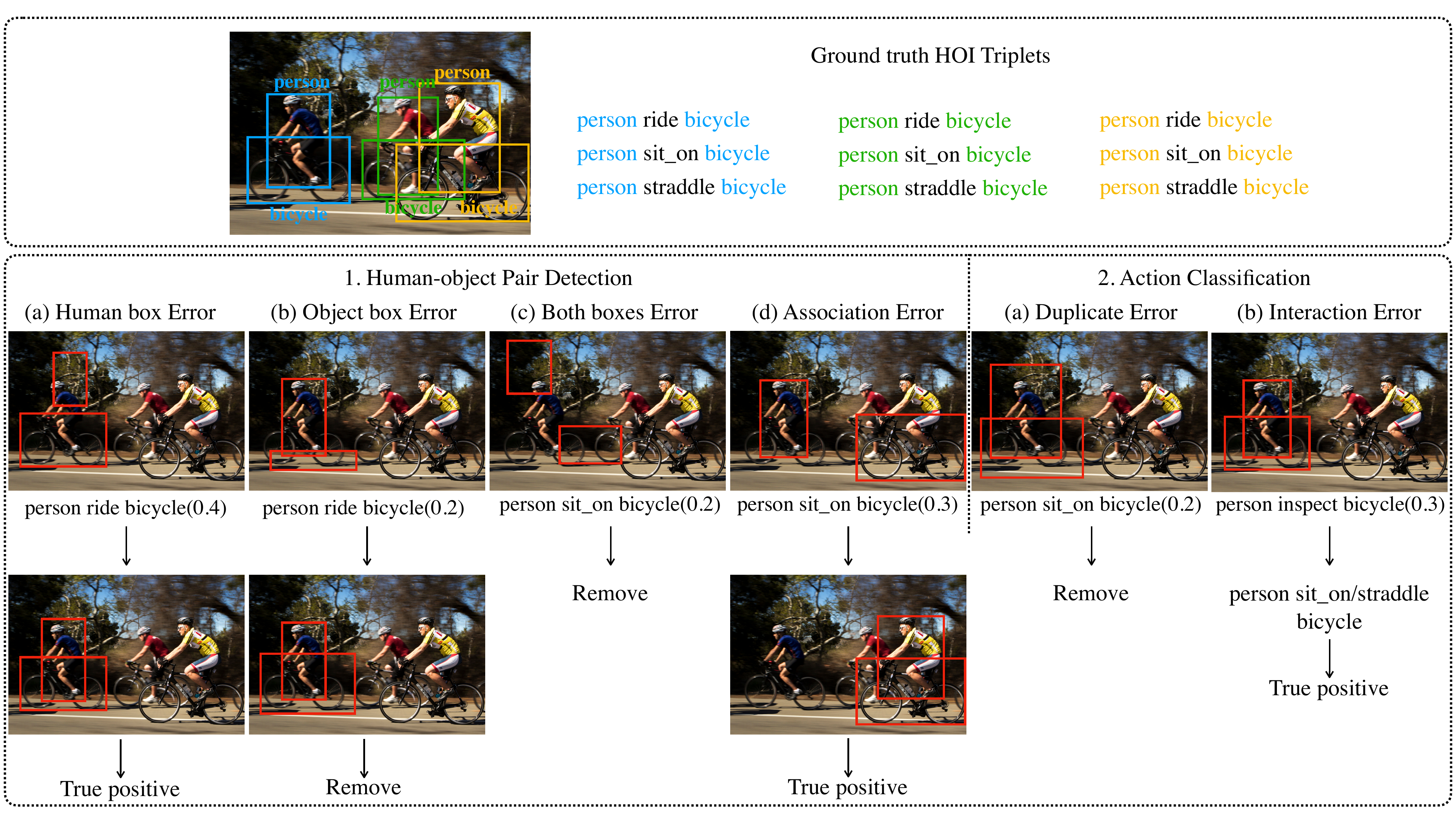}
    \caption{\textbf{Examples of different error types in real images and how we fix them using oracles.} There are three human-object pairs in the ground truth with three interactions between each pair. There are six predicted triplets~(with confidence scores) in the second row, corresponding to six different error types. We fix each of them into true positives and remove duplicates afterwards.}
    \label{fig:fix_errors}
\end{figure*}

\definecolor{gain}{HTML}{34a853}  %
\definecolor{lost}{HTML}{ea4335}  %

\begin{table*}[!htb]
\centering
\caption{\label{tab:recall_hico}\textbf{Average recall and precision of one-stage and two-stage models on HICO-DET dataset.}} 
\renewcommand{\tabcolsep}{6pt}
\newcolumntype{?}{!{\vrule width 1pt}}
\resizebox{0.90\textwidth}{!}{

\begin{tabular}{c|c|c?c|c|c?c|c|c|c|c}
\toprule
\multirow{2}{*}{Arch.} & \multirow{2}{*}{Method} & \multirow{2}{*}{Backbone} & \multicolumn{4}{c|}{Original Models} & \multicolumn{4}{c}{w/ NMS or lower detection threshold $\tau_d$} \\
\cmidrule{4-7}\cmidrule{8-11}
& & & \#Pairs & Pair Rec. & Pair Prec. & HOI $mAP$ & \#Pairs & Pair Rec. & Pair Prec. & HOI $mAP$ \\
\midrule
\multirow{4}{*}{two-stage} & SCG~\cite{zhang2021spatially} & ResNet50 & 44 & 84.0 & 3.7 & 33.3 & \res{78}{34} & \res{86.9}{2.9} & \bad{2.2}{1.5} & \res{33.3}{0.0}\\
 & UPT~\cite{zhang2021efficient} & ResNet50 & 46 & 80.1 & 3.7 & 34.6 & \res{62}{16} & \res{81.7}{1.6} & \bad{2.5}{1.2} & \res{34.8}{0.2} \\
 & UPT~\cite{zhang2021efficient} & ResNet101 & 43 & 80.8 & 3.6 & 35.5 & \res{57}{14} & \res{82.3}{1.5} & \bad{2.8}{0.8} & \res{35.6}{0.1} \\
 & STIP~\cite{zhang2022exploring} & ResNet50 & 32 & 74.2 & 4.4 & 31.6 & \res{64}{32} & \res{75.6}{1.4} & \bad{2.3}{1.9} & \bad{31.1}{0.5} \\
 & RLIPv2~\cite{yuan2023rlipv2} & SwinT & 44 & 80.8 & 3.4 & 41.6 & \res{78}{34} & \res{82.8}{2.0} & \bad{2.5}{0.9} & \res{80.9}{0.1} \\
 & RLIPv2~\cite{yuan2023rlipv2} & SwinL & 45 & 84.0 & 3.5 & 49.1 &
 \res{80}{35} & \res{85.1}{1.1} & \bad{2.5}{1.0} & \res{84.1}{0.1}\\
\midrule
 \multirow{4}{*}{one-stage} & CDN~\cite{zhang2021mining} & ResNet50 & 64 & 79.4 & 2.5 & 34.4 & \bad{26}{38} & \bad{72.0}{7.4} & \res{5.5}{3.0} & \res{34.4}{0.0} \\
 & GEN-VLKT~\cite{liao2022gen} & ResNet50 & 64 & 81.1 & 2.6 & 35.8 & \bad{27}{37} & \bad{78.7}{2.4} & \res{5.5}{1.9} & \res{36.0}{0.2} \\
 & GEN-VLKT~\cite{liao2022gen} & ResNet101 & 64 & 81.8 & 2.6 & 36.5 & \bad{28}{36} & \bad{79.5}{2.3} & \res{5.4}{1.8} & \res{36.5}{0.0} \\
 & QAHOI~\cite{chen2021qahoi} & SwinB & 64 & 81.6 & 1.6 & 35.1 & \bad{45}{19} & \bad{79.7}{1.4} & \res{3.0}{0.7} & \res{35.2}{0.1}\\
 & MUREN~\cite{kim2023relational} & ResNet50 & 64 & 80.1 & 2.5 &  36.0 & \bad{31}{33} & \bad{79.1}{1.0} & \res{3.6}{1.1} & \res{36.1}{0.1}\\
\bottomrule
\end{tabular}

}
\end{table*}

\begin{table*}[t]
\centering
\caption{\label{tab:recall_V-COCO}\textbf{Average recall and precision of one-stage and two-stage models on V-COCO dataset.}} 
\renewcommand{\tabcolsep}{4pt}
\newcolumntype{?}{!{\vrule width 1pt}}
\resizebox{0.90\textwidth}{!}{
\begin{tabular}{c|c|c?c|c|c?c|c|c|c|c}
\toprule
\multirow{2}{*}{Arch.} & \multirow{2}{*}{Method} & \multirow{2}{*}{Backbone} & \multicolumn{4}{c|}{Original Models} & \multicolumn{4}{c}{w/ NMS or lower detection threshold $\tau_d$} \\
\cmidrule{4-7}\cmidrule{8-11}
& & & \#Pairs & Pair Rec. & Pair Prec. & HOI $mAP$ & \#Pairs & Pair Rec. & Pair Prec. & $mAP$ \\
\midrule
\multirow{4}{*}{two-stage} & SCG~\cite{zhang2021spatially} & ResNet50 & 406 & 91.2 & 0.6 & 49.4 & \bad{98}{308} & \bad{90.6}{0.6} & \res{2.5}{1.9} & \res{49.5}{0.1}\\
& UPT~\cite{zhang2021efficient} & ResNet50 & 24 & 92.8 & 10.4 & 59.0 & \res{50}{26} & \res{94.8}{2} & \bad{5.2}{5.2} & \res{59.2}{0.2} \\
& UPT~\cite{zhang2021efficient} & ResNet101 & 26 & 93.2 & 7.8 & 60.7 & \res{55}{29} & \res{94.1}{0.9} & \bad{6.1}{1.7} & \res{60.9}{0.2} \\
 & STIP~\cite{zhang2022exploring} & ResNet50 & 32 & 94 & 7.4 & 67.2 & \bad{23}{9} & \res{92.9}{1.4} & \res{10.4}{3} & \bad{67.1}{0.1} \\
 & RLIPv2~\cite{yuan2023rlipv2} & SwinT & 100 & 93.1 & 2.3 &  66.3 
 & \bad{45}{55} & \bad{92.5}{0.6} & \res{4.0}{1.7} & \res{66.4}{0.1}\\
 & RLIPv2~\cite{yuan2023rlipv2} & SwinL & 100 & 94.0 & 2.4 & 69.5 & \bad{47}{53} & \bad{93.2}{0.8} & \res{3.2}{0.8} & \res{69.5}{0.0}\\
\midrule
\multirow{3}{*}{one-stage} & CDN~\cite{zhang2021mining} & ResNet50 & 100 & 89.7 & 2.3 & 61.7 & \bad{40}{60} & \bad{88.8}{0.9} & \res{5.5}{3.2} & \bad{61.6}{0.1} \\
 & GEN-VLKT~\cite{liao2022gen} & ResNet50 & 64 & 88.5 & 3.5 & 62.4 & \bad{39}{25} & \bad{87.3}{1.2} & \res{5.6}{2.1} & \res{62.4}{0.0} \\
 & GEN-VLKT~\cite{liao2022gen} & ResNet101 & 64 & 89.5 & 3.4 & 63.6 & \bad{37}{27} & \bad{88.1}{1.4} & \res{5.5}{2.1} & \res{63.6}{0.0} \\
 & MUREN~\cite{kim2023relational} & ResNet50 & 100 & 94.0 & 2.4 &  66.2 & \bad{42}{58} & \bad{93.1}{0.9} & \res{3.6}{1.2} & \res{66.3}{0.1}\\
\bottomrule
\end{tabular}
}
\end{table*}

HICO-DET~\cite{chao2018learning} is an extension of the HICO dataset~\cite{chao2015hico} that focuses on image-level HOI classification only. 
In the HICO dataset, 
the \nointeraction label is used to indicate that \emph{none} of the human-object pairs within the image have interactions.
Each of the 80 object categories is associated with a \nointeraction action label, 
resulting in a total of 80 \nointeraction HOI classes. 

HICO-DET inherits the 80 \nointeraction HOI classes in the detection setting, which consists of both localization and classification.
However, it results in inconsistencies between the ground-truth annotations and the evaluation protocol w.r.t. the localization sub-task.
To compute the $mAP$ for the \nointeraction classes, all human-object pairs that have no interactions must be exhaustively annotated. 
Fig.~\ref{fig:no_interaction_example} shows an example of the annotations from the HICO-DET dataset, where we can see the missing annotations for the \nointeraction HOI classes (including both missing bounding boxes and interaction labels). As a result, even if a model correctly outputs the \nointeraction labels,
they will be considered as false positives and the model's $mAP$ gets penalized. 
Note this is not an issue for the HOI \emph{classification} in HICO~\cite{chao2015hico} as no localization is needed.

How can we solve this issue? Obviously, annotating all the missing \nointeraction human-object pairs exhaustively is not feasible nor scalable. 
In fact, the \nointeraction HOI category is not needed. %
If there are no annotations stating that two objects have any \emph{actual} interactions (\eg, \texttt{catch} or \texttt{ride}), it means they have no interactions, as what the current annotations indicate in Fig.~\ref{fig:no_interaction_example}.
This setting is adopted in the V-COCO~\cite{gupta2015visual} benchmark.
Therefore, in our diagnosis, we remove all 80 \nointeraction HOI classes and only consider the remaining 520 ones for the HICO-DET benchmark~\cite{chao2018learning}.

We would like to emphasize that not computing the $mAP$ for the \nointeraction HOI category does not change an HOI detector's output to ignore the human-object pairs that have no interactions nor do we underestimate the detector's accuracy in our diagnosis. First, we do not remove the \nointeraction label in the model's output so there is no need to re-train the model. 
An HOI model is still able to tell that a human-object pair has no interactions, which we discuss in detail in Section~3.3.
Second, if the model incorrectly classifies a human-object pair that has no interaction as having an actual interaction (\eg, \texttt{ride bicycle}), such an incorrect output will be considered as a false positive. Similarly, incorrectly classifying the pair of \texttt{ride bicycle} as \nointeraction will reduce the number of true positives. Both of such errors will lead to a lower $mAP$ that correctly measures the model's accuracy.

\section{Details of fixing errors using oracles}
\label{sec:oracles}
In this section, we provide details of how we fix errors according to different oracles. We show visualization examples of how we fix different types of errors in Fig.~\ref{fig:fix_errors}.

\noindent\textbf{Fixing the human-object pair detection errors.} 
It involves four oracles.
\begin{itemize}
    \item \textbf{Human box oracle}: 
    Fix the human box detection and action label, making it a true positive. If duplicates are made, suppress the lower-scoring prediction. 
    \item \textbf{Object box oracle}: Fix such false positives in a similar way as introduced in the previous step.
    \item \textbf{Both boxes oracle}: Since both boxes are incorrect, we cannot decide which ground truth triplet the detection is attempting to match. We just remove this kind of false positive prediction.
    
    \item \textbf{Association oracle}: 
    Correct the pair association and action label, making it a true positive. If duplicates are made in this way, suppress the lower-scoring prediction.
\end{itemize}

\noindent\textbf{Fixing interaction classification errors.} 
In this case, we fix false positives caused by duplicate error or interaction error.
\begin{itemize}
    \item \textbf{Duplicate oracle}: We directly remove all duplicate predictions.
    \item \textbf{Interaction classification oracle}: Fix the action label to make it a true positive. If duplicates are made in this way, suppress the lower-scoring prediction. 
\end{itemize}

\noindent\textbf{Fixing missed GT errors.}
Note that, after fixing all the above errors, there are still ground truth triplets that are not matched with any predictions, the number of which is the number of missed ground truth triplets.
We reduce the number of ground-truth HOI triplets in the $mAP$ calculation by the number of missed ground-truth triplets.

\noindent\textbf{Grouping the errors.}
\begin{itemize}
    \item \textbf{False positive oracle}: Remove all false positive predictions.
    \item \textbf{False negative oracle}: Set the number of ground truth triplets to the number of true positive predictions.
\end{itemize}

\noindent\textbf{Oddities of $mAP$ improvement.}
Similar to \cite{bolya2020tide}, the $mAP$ improvement in our case has the same issue that the summation of $\Delta mAP$ of different error types does not lead to $100-mAP$. For example, in Fig.~5 of the main paper, adding $mAP$ of CDN~\cite{zhang2021mining} with $\Delta mAP_{FP}$ and $\Delta mAP_{FN}$ (34.4+33.1+12.94) yields 80.44, not 100. As pointed out in \cite{bolya2020tide}, the reason is that fixing different errors at once gives a larger $mAP$ improvement than fixing each error on its own.

\section{Additional pair localization results}

As introduced in Section~2.3, the NMS is applied before the pairing of human and object boxes in two-stage methods, which explains why the number of detected pairs of one-stage methods is higher than the two-stage counterparts.
To examine the impact of this factor, we apply NMS to remove duplicate human-object pair detections for one-stage methods, and lower the detection threshold to increase the human-object pair detections for two-stage methods. After that their numbers of pairs become comparable, and we recalculate Pair Recall, Pair Precision and HOI detection $mAP$.
Note that, the two-stage RLIPv2~\cite{yuan2023rlipv2} does not apply NMS in their original model, so we use NMS to reduce its number of pairs for comparison.

As shown in the right parts of Tab.~\ref{tab:recall_hico} and \ref{tab:recall_V-COCO}, larger number of pairs (either because of no NMS or lower detection threshold) leads to higher Pair Recall and lower Pair Precision.
However, neither increasing nor decreasing the number of pairs leads to a significant enhancement of the final HOI detection $mAP$. 

\begin{figure*}[t]
    \centering
    \includegraphics[width=1.0\textwidth]{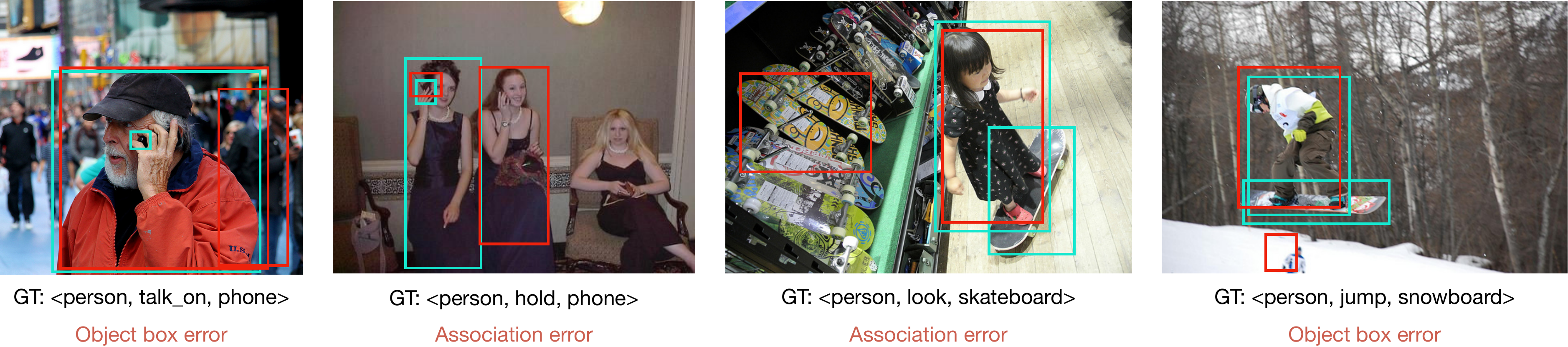}
    \caption{\textbf{Visualization of errors of SCG~\cite{zhang2021spatially} and UPT~\cite{zhang2021efficient} on the V-COCO dataset.} We show the ground truth triplet (in cyan color) and the error type of the prediction~(red characters) for each image. The red boxes represent predictions.}
    \label{fig:vis_V-COCO_preds}
\end{figure*}

\section{Models Used in Our Diagnosis}
We choose eight popular HOI detection models including three one-stage and three two-stage detectors for our diagnosis. 
We give a brief summary for each model we used.

\noindent\textbf{SCG~\cite{zhang2021spatially}} solves the HOI detection problem by graphical neural networks with a two-stage design. It designs condition messages between pairs of nodes on their spatial relationships.

\noindent\textbf{UPT~\cite{zhang2021efficient}}
proposes an Unary–Pairwise Transformer architecture, but makes it a two-stage model that exploits unary and pairwise representations for
HOIs. 

\noindent\textbf{STIP~\cite{zhang2022exploring}} is a two-stage method that 
uses a Transformer-based detector to generate interaction proposals first, and then transforms the nonparametric interaction proposals into HOI predictions via a structure-aware Transformer.

\noindent\textbf{RLIPv2~\cite{yuan2023rlipv2}} is a two-stage,
fast converging model, enabling large-scaled relational pre-training with pseudo-labelled data. After pre-training, it performs well on both HOI detection and scene graph generation.

\noindent\textbf{CDN~\cite{zhang2021mining}} proposes a one-stage method with disentangling
human-object detection and interaction classification in a cascade manner. 
It first uses a human-object pair generator and then designs an isolated interaction classifier to classify each human-object pair.

\noindent\textbf{QAHOI~\cite{chen2021qahoi}} proposes a transformer-based one-stage 
method, which leverages a multi-scale architecture to extract features from different scales and uses query-based anchors to predict human-object pairs and their interactions as triplets.

\noindent\textbf{GEN-VLKT~\cite{liao2022gen}} follows the one-stage cascaded manner of CDN, and designs guided embeddings and instance guided embeddings to generate HOI instances.
Besides, it proposes a Visual-Linguistic Knowledge Transfer
training strategy for better interaction understanding by
transferring knowledge from a pre-trained model CLIP~\cite{radford2021learning}.

\noindent\textbf{MUREN~\cite{kim2023relational}} follows the one-stage paradigm and designs three decoder branches using unary, pairwise, and ternary relations of human, object, and interaction tokens for discovering HOI instances.

\section{Visualization of errors on V-COCO}
We randomly choose test images from the V-COCO dataset to see the predictions of SCG~\cite{zhang2021spatially} and UPT~\cite{zhang2021efficient}, as shown in Fig~\ref{fig:vis_V-COCO_preds}.  Since these two methods use the postprocessing method, there are no action errors in the predictions.

\section{Human Detectioin}
Previous analysis of $mAP$ improvement shows that the human error is not as significant as the generic objects. 
Here, we examine the recall of human detection. On the HICO-DET dataset, the average recall of human detection is 91.5, while the average recall of all object categories is 88.0\footnote{Since the annotations of objects are not complete as we pointed out in Section 3.1 in the main paper, the precision is not reliable.}. 
We can see that human detection is easier but the performance is still far from satisfactory.

{
    \small
    \bibliographystyle{ieeenat_fullname}
    \bibliography{main}
}

\end{document}